\documentclass{article}

\usepackage[scaled=0.85]{DejaVuSansMono}

\renewcommand{\ttfamily}{\fontfamily{DejaVuSansMono-TLF}\selectfont}

\renewcommand{\texttt}[1]{{\fontfamily{DejaVuSansMono-TLF}\selectfont #1}}

\usepackage{minted} 
\setminted{
    fontfamily=DejaVuSansMono-TLF,
    fontsize=\small
}

\usepackage[preprint]{neurips_2026}

\usepackage[utf8]{inputenc} 
\usepackage[T1]{fontenc}    
\usepackage{url}            
\usepackage{booktabs}       
\usepackage{amsfonts}       
\usepackage{nicefrac}       
\usepackage{microtype}      
\usepackage[table]{xcolor}  
\usepackage{lipsum}
\usepackage{adjustbox}
\usepackage{pgfplotstable}
\usepackage{pifont}         
\newcommand{\cmark}{\ding{51}}
\newcommand{\xmark}{\ding{55}}

\usepackage{amsmath}
\usepackage{amssymb}
\usepackage{mathtools}
\usepackage{enumitem}

\usepackage{amsthm}

\usepackage{multirow}       

\usepackage{url}
\usepackage[utf8]{inputenc} 
\usepackage[T1]{fontenc}    
\usepackage[breaklinks]{hyperref}      

\usepackage{booktabs}       
\usepackage{amsfonts}       
\usepackage{amsmath}
\usepackage[mathscr]{euscript}
\usepackage{amssymb}
\usepackage{mathtools}
\usepackage{amsthm}
\usepackage{nicefrac}       
\usepackage{microtype}      

\usepackage{graphicx}
\usepackage{xcolor}      

\usepackage{caption}
\usepackage{multirow,makecell}

\usepackage{xspace}
\usepackage{subcaption}

\usepackage{algorithm}
\usepackage{algorithmic}
\usepackage{wrapfig}

\usepackage[most]{tcolorbox}

\usepackage[capitalize,noabbrev]{cleveref}

\title{Talking Trees: Reasoning-Assisted Induction of \\ Decision Trees for Tabular Data}

\author{%
    George Yakushev$^*$ \\ Yandex, HSE University \\ \And
    Alina Shutova$^*$ \\ Yandex, HSE University \\ \And
    Ivan Rubachev$^*$ \\ Yandex, HSE University \\ \And
    Natalia Bereberdina$^*$ \\ Yandex, HSE University \\ \And
    Renat Sergazinov$^*$ \\ Yandex \\ \And
    Artem Babenko$^\dagger$ \\ Yandex, HSE University \\ 
}

\begin{document}

\maketitle

\begin{abstract}
Tabular foundation models are becoming increasingly popular for low-resource tabular problems. These models make up for small training datasets by pretraining on large volumes of synthetic data. The prior knowledge obtained via pretraining provides the exceptional performance, but the resulting model becomes a black box that is difficult to interpret and costly for inference.
In this work, we explore an alternative strategy: using reasoning-capable LLMs to induce decision trees for small tabular datasets in an agentic setup. We design a minimal set of tools for constructing, analyzing, and manipulating decision trees. Equipped with these tools, the LLM combines its prior knowledge with learning from data to produce a lightweight decision tree that outperforms CART and recent non-greedy tree learners and remains competitive with tree ensembles on low-resource tabular problems.
While a single agentic decision tree is competitive with state-of-the-art black box models, it also comes with a human-readable reasoning trace that can be checked for biases and data leaks. Furthermore, the reasoning-based LLM's creation process allows for additional human input to be incorporated into the tree without it being captured in data.
\end{abstract}

\vspace{-10pt}
\section{Introduction}\label{sect:intro}
\vspace{-8pt}

Tabular prediction problems are often treated as learning from a self-contained table of samples. In practice, many real deployments come with \emph{extra context} that is only partially reflected in the dataset: domain meta-knowledge (e.g., known causal directions or data artifacts), constraints on how predictions should behave (e.g., fairness, safety, monotonicity), and application-specific rules imposed by regulation. When data are limited, this missing context can matter as much as the statistical signal.

Recent progress on small/medium tabular benchmarks is dominated by in-context learning \emph{tabular foundation models}~\cite{hollmann2022tabpfn, zhang2025limix, jingangtabicl, rubachev2025finetuning, liu2025tabpfn}, which are pretrained on large collections of synthetic or real tabular tasks and achieve strong few-shot performance, often surpassing classical tree ensembles such as Random Forests~\cite{breiman2001random} and GBDTs~\cite{friedman2001greedy,chen2016xgboost}. However, for practitioners these models remain largely \emph{black boxes}: their decision rules are implicit in weights and inference procedures, which limits interpretability and verifiability. Moreover, they can be costly at deployment time, requiring substantial memory and compute per prediction~\cite{erickson2025tabarena}. Using LLMs as inference engines or model boosters has been explored in \citep{hegselmann2023tabllm, gardner2024large, nam2024optimized, ye2025_llm_meeting_decision_trees}, but these approaches typically suffer from high inference costs or remain black boxes.

In this work, we explore a different point in the design space: we use a reasoning-capable LLM agent \emph{only at training time} to construct a \emph{standalone} decision tree model capable of accommodating arbitrary user constraints. Concretely, we formulate tree induction as an agentic thought--action--observation loop and equip the LLM with a small toolkit for decision-tree construction and refinement: data preprocessing, split search, validation, pruning, and targeted subtree edits. Modern LLMs such as GPT-5~\cite{gpt5}, DeepSeek R1~\cite{guo2025deepseek}, and GLM-4.5~\cite{zeng2025glm} can already use these tools to iteratively improve a tree, combining learning from data with prior knowledge. The resulting predictors are competitive with state-of-the-art models in the low-resource setting while remaining fully interpretable, and outperform both greedy CART and recent non-greedy tree learners such as TAO~\cite{carreira2018alternating}.

\begin{figure*}[t]
    \centering
    \vspace{-5px}
    \includegraphics[width=\textwidth]{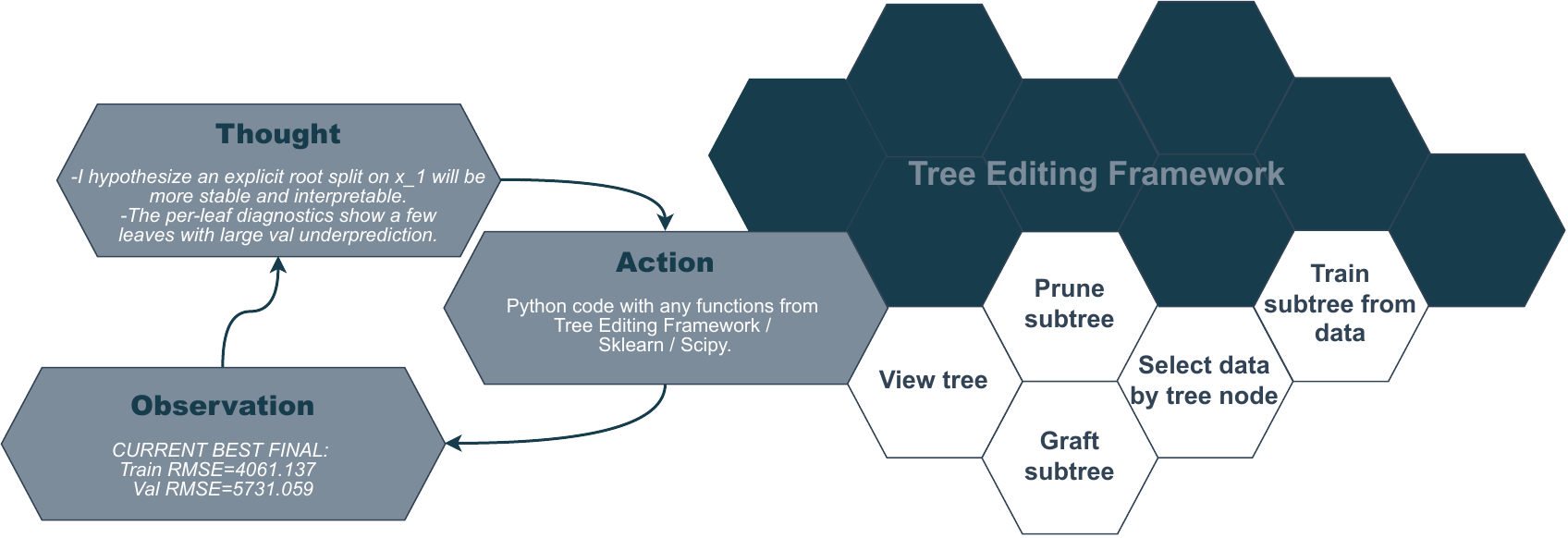}
    \vspace{-10px}
    \caption{Informal overview. We prompt an LLM agent to construct a decision tree in a thought--action--observation cycle (left). During the action phase, the agent uses a tree-editing framework (right) with tools for analyzing and modifying trees. The agent can combine these tools in Python and print arbitrary diagnostics for the observation phase.}
    \label{fig:teaser}
    \vspace{-10px}
\end{figure*}

\textbf{Lightweight inference.}
A key benefit of this setup is that inference is extremely cheap: once the tree is constructed, predictions require \emph{no LLM calls} and reduce to a small sequence of feature comparisons (the same deployment footprint as a standard decision tree). The cost is paid once during training, making it feasible to use strong LLMs without incurring per-example inference overhead.

\textbf{Controllability.}
Because the tree is produced through tool-augmented reasoning, the learning objective can be specified \emph{informally} in the prompt, enabling inclusion of arbitrary loosely-defined constraints. We study several examples where this flexibility is useful in Section~\ref{sec:control}: (1) requesting the tree to reduce dependence on protected attributes (fairness), (2) incorporating a feature that will exist at deployment time but is absent from historical training data via a natural-language description (missing feature), and (3) enforcing domain constraints such as monotonic relationships or other business rules. These capabilities are complementary to pure performance.

\textbf{Interpretability.}
Our approach yields an explicit \emph{construction trace} (tool calls used and intermediate trees) that can be inspected to diagnose failures, detect undesirable shortcuts, and support governance in high-stakes settings~\cite{korbak2025_cot_monitorability,Arnav2025CoTRS}. While such traces are not a formal guarantee, they provide a practical handle for debugging and verification that is largely absent in black-box tabular predictors. 

Our contributions are as follows:
\begin{itemize}[leftmargin=*]
\vspace{-4px}\item We introduce a training-time procedure that uses a reasoning-capable LLM and a small toolkit to construct, verify, and refine decision trees for tabular tasks, producing a \emph{standalone} model with \emph{lightweight inference}.
\vspace{-4px}\item We show that tool-augmented LLMs construct trees that outperform CART decision trees and remain competitive with recent non-greedy tree learners, while providing an interpretable alternative to black-box tabular foundation models that is controllable through prompt-level requirements.
\vspace{-4px}\item We release the evaluation code\footnote{\url{https://github.com/yandex-research/TalkingTrees}} and the benchmarking tools in supplementary materials
to facilitate comparisons with future LLMs.
\end{itemize}

\vspace{-10pt}
\section{Related Work}\label{sect:background}
\vspace{-8pt}

Prior to our work, there have been a number of attempts to develop foundational models for tabular data with LLMs. For convenience, we organize these attempts into two groups: (1) inference-time in-context learning and (2) training-time tuning for zero-shot prediction. Table~\ref{tab:agentic-comparison} summarizes representative approaches and highlights key trade-offs.
\begin{table}[t]
\centering
\vspace{-10pt}
\caption{\textbf{Comparison} of LLM-based methods. $^\dagger$Interpretability mainly at the \emph{feature} level.}
\scriptsize
\begin{tabular}{l l c c c}
\toprule
\textbf{Method} & \textbf{Family} & \textbf{Steerable} & \textbf{Interpretable} & \textbf{Light Inference} \\
\midrule
\rowcolor{blue!8}
\textbf{Agentic Tree (ours)} & Agentic Tree & \cmark & \cmark & \cmark \\
\midrule
OCTree~\cite{nam2024optimized} & FE Agent & \xmark & \cmark$^\dagger$ & \cmark \\
CAAFE~\cite{hollmann2023_caafe} & FE Agent & \xmark & \cmark$^\dagger$ & \cmark \\
LLM-FE~\cite{abhyankar2025llm} & FE Agent & \xmark & \cmark$^\dagger$ & \cmark \\
DeLTA~\cite{ye2025_llm_meeting_decision_trees} & Booster & \xmark & \xmark & \cmark \\
TabLLM~\cite{hegselmann2023tabllm} & LLM & \cmark & \xmark & \xmark \\
TabuLa-8B~\cite{gardner2024large} & LLM & \cmark & \xmark & \xmark \\
TabPFN~\cite{hollmann2022tabpfn} & ICL & \xmark & \xmark & \xmark \\
TabICL~\cite{jingangtabicl} / LimiX~\cite{zhang2025limix} & ICL & \xmark & \xmark & \xmark \\
\bottomrule
\end{tabular}
\label{tab:agentic-comparison}
\vspace{-10pt}
\end{table}

\textbf{Inference time.} \nocite{brown2020language} It has been shown that LLMs pre-trained on vast amounts of data can solve tabular problems at inference time~\cite{lovre2022_who_models_models}. This line includes direct prompting approaches such as TabLLM~\cite{hegselmann2023tabllm} and TabuLa-8B~\cite{gardner2024large}, which are typically steerable but require an LLM forward pass for each prediction. Subsequent works explore this further by ensembling in-context-learned LLMs~\cite{manikandan2023_language_models_are_weak_learners} and using LLM reasoning on top of traditional models~\cite{liu2025_chain_of_thoughts_tabular}. Recent works in this field also propose ways to enhance the LLM's ability to solve tabular problems with specialized fine-tuning~\cite{gardner2024large}.\nocite{LLMDATASurvey2025}
More broadly, foundational ICL-style tabular models such as TabPFN~\cite{hollmann2022tabpfn}, TabICL~\cite{jingangtabicl}, and LimiX~\cite{zhang2025limix} amortize training into a learned prior and can offer lighter per-example inference, but still rely on in-context learning and generally provide limited end-to-end interpretability.

\textbf{Training time.}
Another recent line of work instead proposes to use LLMs in the model learning phase. In the simplest case, LLMs can be used for automated feature engineering, e.g., CAAFE~\cite{hollmann2023_caafe}, OCTree~\cite{nam2024optimized}, and LLM-FE~\cite{abhyankar2025llm}, and for data cleaning~\cite{zhang2024_data_cleaning_using_llms,li2024_autodcworkflow,bendinelli2025_exploring_llm_agents_cleaning}. More recent works even use LLM-based agents to automate top-level machine learning pipelines~\cite{jiang2025_aide,fang2025_mlzero}, including writing code and tuning hyperparameters.
The research line most relevant to our work investigates LLM-based methods that interact directly with decision trees~\cite{carrasco2025zero,knauer2025oh,ye2025_llm_meeting_decision_trees}. Unlike more general machine learning models (e.g., MLPs), a decision tree has an explicit logical structure that can be analyzed and generated by LLMs. The first two works~\cite{carrasco2025zero,knauer2025oh} leverage LLMs to construct a decision tree from the problem description in a zero-shot fashion. DeLTA~\cite{ye2025_llm_meeting_decision_trees} uses an LLM in a zero-shot setting to edit the decision rules of a random forest and then fits an error-correcting layer based on either TabPFN~\cite{hollmann2022tabpfn} or CART.
In this work, we consider a more general setup: rather than tasking the LLM to follow a rigid procedure, we use it in an agentic mode, equipping it with a set of tools to iteratively build, analyze, and refine a decision tree using both data and domain knowledge. This yields a lightweight and interpretable decision tree model.

\textbf{Optimal decision trees.} A complementary line of research seeks to overcome the suboptimality of greedy CART-style induction by formulating tree learning as a global optimization problem. Mixed-integer optimization formulations~\cite{bertsimas2017optimal} produce provably optimal trees but scale poorly with dataset size and depth, motivating alternative encodings based on SAT solvers~\cite{narodytska2018sat}. To trade exactness for tractability, TAO~\cite{carreira2018alternating} alternates over disjoint subsets of nodes to directly minimize the misclassification loss, while DPDT~\cite{kohler2025breiman} casts tree induction as a Markov decision process with heuristic split generation, bridging the greedy and optimal regimes. These methods consistently improve over CART when the training objective aligns with the true predictor, but they cannot learn non-linear dependencies, accommodate fairness constraints, or incorporate prior knowledge about the data. Moreover, both DPDT and SAT currently produce only hard labels for classification, and SAT does not support regression at all.

\vspace{-10pt}
\section{Agentic Tree Induction}\label{sect:method}
\vspace{-8pt}

We reformulate the process of learning a decision tree as an agentic task~\cite{Schick2023ToolformerLM,Qin2023ToolLLMFL}, giving the LLM tools to expand, analyze, and manipulate the decision tree. The LLM agent does not generate the entire tree in one go, but gradually refines it in small increments. We hypothesize that this incremental process will allow LLMs to better harness their capacity for chain-of-thought reasoning by testing their hypotheses against data and correcting their own errors\nocite{cot_wei_2022,zero_shot_cot_Kojima2022LargeLM,auto_cot_Zhang2022AutomaticCO}. To facilitate model building, we design a set of tools for tree analysis and manipulation. These tools are organized as a single-file\footnote{See \texttt{editable\_tree.py} in the supplementary code.}.
\begin{wrapfigure}{r}{0.52\textwidth}
\vspace{-12px}
\begin{minted}{python}
Tree("feat1<=0.5",       # if feat1 <= 0.5:
     le=3.5,             #   return 3.5
     gt=Tree("feat2<=9", # elif feat2 <= 9:
             le=-2,      #   return -2
             gt=+2))     # else: return 2
\end{minted}
\vspace{-15px}
\end{wrapfigure}
Python library designed around the \texttt{Tree} class. The tree can be converted from a \texttt{Scikit-Learn} tree\footnote{Default \texttt{sklearn} trees are hard to edit.}~\cite{scikit-learn} or created via the code:\nocite{python}

We give the LLM agent access to the following tools for analyzing and modifying the tree:
\begin{enumerate}[leftmargin=*]
    \item \textbf{View tree:} \texttt{print(tree)} displays the tree or chosen subtree structure as text. The printed tree includes node IDs that can be used via \texttt{tree.find\_node(id)}.
    \item \textbf{Prune subtree:} converts a tree node into a leaf. This can be done manually, e.g., \texttt{tree.gt.prune()}, or by node ID via \texttt{tree.find\_node(id).prune()}.
    \item \textbf{Select data:} run the tree on dataset \texttt{X} and select samples that pass through a given node ID: \texttt{tree.get\_data\_indices\_for\_node(id, X)}.
    \item \textbf{Graft subtree:} replace the specified node (by ID) in the current tree with a new (manual or learned) subtree: \texttt{tree.replace\_subtree(id, new\_subtree)}.
\end{enumerate}

Tools were designed for the agent to perform more complex tree manipulations, as shown on the example:

\begin{wrapfigure}{l}{0.5\textwidth}
\vspace{-15px}
\begin{minted}{python}
sktree = DecisionTreeRegressor(max_depth=6)
sktree.fit(X_train, y_train)
tree = Tree.from_sklearn(sktree)
print(tree)  # observe node IDs
tree.find_node(42).prune()
# manual tree editing
node = tree.find_node(43)
node.feature = "sepal_length"
node.threshold = 6.5
node.gt = Tree("width<=5", le=-3, gt=+3)
node.le = Tree(value=0) # leaf
# semi-automated manipulation
ix = tree.get_data_indices_for_node(
    id=44, X=X_train
)
features = my_features(X_train.iloc[ix])
subtree = DecisionTreeRegressor(max_depth=2)
subtree.fit(
    X=X_train.iloc[ix][features],
    y=y_train[ix]
)
tree.replace_subtree(
    44, Tree.from_sklearn(subtree)
)
\end{minted}
\vspace{-15px}
\end{wrapfigure}

Our framework also implements several smaller utility tools, such as accessing node properties (feature, threshold), re-indexing node IDs after edits, and refitting tree leaves from data, all included in the supplementary code.
In addition to these tools, the Python-based LLM agent has access to NumPy~\cite{harris2020arraynumpy}, SciPy~\cite{virtanen2020scipy}, pandas~\cite{reback2020pandas}, Scikit-Learn~\cite{scikit-learn}, and common Python builtins such as \texttt{math}.

We implement the agent using \texttt{smolagents.CodeAgent}~\cite{smolagents}, with a custom environment and system prompt. We pre-load the data, the allowed packages, and the tools into the agent's environment. The system prompt\footnote{The full system prompt is included in the supplementary code.} then outlines the task and the environment setup. When building a tree for a tabular dataset, the agent receives the training and validation subsets, \textit{but it has no access to the test data}. Additionally, we provide the LLM with a short description of the input features and the objective. Aside from that, we use the default \texttt{smolagents} hyperparameter of up to 20 thought--code--observation cycles. The entire process takes ${\le}10$ minutes and ${\le}0.3\$$ in API costs per tree even for state-of-the-art LLMs (e.g., GPT-5) and can be parallelized for ensembling.

\textbf{Additional human input.} Using general-purpose LLM agents comes with safety concerns, but it also introduces an important new capability: harnessing informal domain knowledge (as prompt) in addition to the training data. This covers a broad range of use cases, from explaining the underlying physical model for engineering problems to sharing intuition or adjusting for known problems in the data. In Section~\ref{sec:control}, we explore multiple such cases: correcting for a known bias, using a new feature, and imposing monotonic constraints on the model.

\begin{table*}[t]
\centering
\caption{Evaluating agent-constructed decision trees and advanced tabular methods on low-resource tabular datasets. Columns are grouped by task type, while rows separate single-tree methods from other ensemble, hybrid, or black-box methods. \textcolor{gray}{Gray} values are standard errors over 5 splits (TabArena v0.1). \textbf{Bold} indicates the best result within each row block.}
\label{tab:main_table}

\small
\setlength{\tabcolsep}{2pt}
\renewcommand{\arraystretch}{1.2}
\begin{adjustbox}{max width=\textwidth}
\begin{tabular}{l*{17}{c}}
\toprule
 &
\multicolumn{8}{c}{Classification: Binary (ROC AUC$\uparrow$)} &
\multicolumn{4}{c}{Multiclass (LogLoss$\downarrow$)} &
\multicolumn{5}{c}{Regression (RMSE$\downarrow$)} \\
\cmidrule(lr){2-9}\cmidrule(lr){10-13}\cmidrule(lr){14-18}
& \makecell{Fitness}
& \makecell{Marketing}
& \makecell{CreditG}
& \makecell{Diabetes}
& \makecell{Customer}
& \makecell{QSARBio}
& \makecell{Hazelnut}
& \makecell{Blood}
& \makecell{Anneal}
& \makecell{Maternal}
& \makecell{Phishing}
& \makecell{MIC}
& \makecell{Airfoil}
& \makecell{Fiat500}
& \makecell{Concrete}
& \makecell{Insurance}
& \makecell{QSARFish} \\
\midrule

\multicolumn{18}{l}{\textit{\textbf{Single-tree methods}}} \\
\makecell[l]{CART [Default]\\[2pt]}       & 
\shortstack{\(0.615\)\\{\tiny\textcolor{gray}{\(\pm 0.007\)}}} &
\shortstack{\(0.672\)\\{\tiny\textcolor{gray}{\(\pm 0.009\)}}} &
\shortstack{\(0.615\)\\{\tiny\textcolor{gray}{\(\pm 0.013\)}}} &
\shortstack{\(0.676\)\\{\tiny\textcolor{gray}{\(\pm 0.017\)}}} &
\shortstack{\(0.549\)\\{\tiny\textcolor{gray}{\(\pm 0.008\)}}} &
\shortstack{\(0.781\)\\{\tiny\textcolor{gray}{\(\pm 0.008\)}}} &
\shortstack{\(0.818\)\\{\tiny\textcolor{gray}{\(\pm 0.008\)}}} &
\shortstack{\(0.554\)\\{\tiny\textcolor{gray}{\(\pm 0.018\)}}} &
\shortstack{\(0.529\)\\{\tiny\textcolor{gray}{\(\pm 0.124\)}}} &
\shortstack{\(5.44\)\\{\tiny\textcolor{gray}{\(\pm 0.58\)}}} &
\shortstack{\(3.82\)\\{\tiny\textcolor{gray}{\(\pm 0.19\)}}} &
\shortstack{\(6.89\)\\{\tiny\textcolor{gray}{\(\pm 0.24\)}}} &
\shortstack{\(2.957\)\\{\tiny\textcolor{gray}{\(\pm 0.044\)}}} &
\shortstack{\(1003.1\)\\{\tiny\textcolor{gray}{\(\pm 6.64\)}}} &
\shortstack{\(7.67\)\\{\tiny\textcolor{gray}{\(\pm 0.13\)}}} &
\shortstack{\(6380.9\)\\{\tiny\textcolor{gray}{\(\pm 191.1\)}}} &
\shortstack{\(1.299\)\\{\tiny\textcolor{gray}{\(\pm 0.015\)}}} \\

\makecell[l]{CART [Tuned]\\[2pt]}         & 
\shortstack{\(0.800\)\\{\tiny\textcolor{gray}{\(\pm 0.009\)}}} &
\shortstack{\(0.799\)\\{\tiny\textcolor{gray}{\(\pm 0.012\)}}} &
\shortstack{\(0.700\)\\{\tiny\textcolor{gray}{\(\pm 0.015\)}}} &
\shortstack{\(0.779\)\\{\tiny\textcolor{gray}{\(\pm 0.012\)}}} &
\shortstack{\(0.698\)\\{\tiny\textcolor{gray}{\(\pm 0.005\)}}} &
\shortstack{\(0.852\)\\{\tiny\textcolor{gray}{\(\pm 0.011\)}}} &
\shortstack{\(0.895\)\\{\tiny\textcolor{gray}{\(\pm 0.006\)}}} &
\shortstack{\(0.698\)\\{\tiny\textcolor{gray}{\(\pm 0.013\)}}} &
\shortstack{\(0.321\)\\{\tiny\textcolor{gray}{\(\pm 0.030\)}}} &
\shortstack{\(0.849\)\\{\tiny\textcolor{gray}{\(\pm 0.063\)}}} &
\shortstack{\(0.674\)\\{\tiny\textcolor{gray}{\(\pm 0.088\)}}} &
\shortstack{\(0.538\)\\{\tiny\textcolor{gray}{\(\pm 0.011\)}}} &
\shortstack{\(2.988\)\\{\tiny\textcolor{gray}{\(\pm 0.060\)}}} &
\shortstack{\(\mathbf{789.8}\)\\{\tiny\textcolor{gray}{\(\pm 14.13\)}}} &
\shortstack{\(7.65\)\\{\tiny\textcolor{gray}{\(\pm 0.17\)}}} &
\shortstack{\(4494.3\)\\{\tiny\textcolor{gray}{\(\pm 74.8\)}}} &
\shortstack{\(1.043\)\\{\tiny\textcolor{gray}{\(\pm 0.021\)}}} \\

\makecell[l]{TAO\\[2pt]} &
\shortstack{\(0.806\)\\{\tiny\textcolor{gray}{\(\pm 0.006\)}}} &
\shortstack{\(0.812\)\\{\tiny\textcolor{gray}{\(\pm 0.008\)}}} &
\shortstack{\(0.677\)\\{\tiny\textcolor{gray}{\(\pm 0.011\)}}} &
\shortstack{\(0.778\)\\{\tiny\textcolor{gray}{\(\pm 0.007\)}}} &
\shortstack{\(0.650\)\\{\tiny\textcolor{gray}{\(\pm 0.021\)}}} &
\shortstack{\(0.839\)\\{\tiny\textcolor{gray}{\(\pm 0.012\)}}} &
\shortstack{\(0.859\)\\{\tiny\textcolor{gray}{\(\pm 0.020\)}}} &
\shortstack{\(\mathbf{0.774}\)\\{\tiny\textcolor{gray}{\(\pm 0.009\)}}} &
\shortstack{\(0.200\)\\{\tiny\textcolor{gray}{\(\pm 0.039\)}}} &
\shortstack{\(0.936\)\\{\tiny\textcolor{gray}{\(\pm 0.101\)}}} &
\shortstack{\(1.341\)\\{\tiny\textcolor{gray}{\(\pm 0.323\)}}} &
\shortstack{\(0.588\)\\{\tiny\textcolor{gray}{\(\pm 0.028\)}}} &
\shortstack{\(3.005\)\\{\tiny\textcolor{gray}{\(\pm 0.060\)}}} &
\shortstack{\(799.1\)\\{\tiny\textcolor{gray}{\(\pm 8.45\)}}} &
\shortstack{\(7.547\)\\{\tiny\textcolor{gray}{\(\pm 0.146\)}}} &
\shortstack{\(\mathbf{4409.6}\)\\{\tiny\textcolor{gray}{\(\pm 58.8\)}}} &
\shortstack{\(1.028\)\\{\tiny\textcolor{gray}{\(\pm 0.033\)}}} \\

\makecell[l]{Ours [CART]\\[2pt]}   & 
\shortstack{\(\mathbf{0.818}\)\\{\tiny\textcolor{gray}{\(\pm 0.008\)}}} &
\shortstack{\(\mathbf{0.820}\)\\{\tiny\textcolor{gray}{\(\pm 0.014\)}}} &
\shortstack{\(\mathbf{0.717}\)\\{\tiny\textcolor{gray}{\(\pm 0.005\)}}} &
\shortstack{\(\mathbf{0.799}\)\\{\tiny\textcolor{gray}{\(\pm 0.013\)}}} &
\shortstack{\(\mathbf{0.717}\)\\{\tiny\textcolor{gray}{\(\pm 0.008\)}}} &
\shortstack{\(\mathbf{0.854}\)\\{\tiny\textcolor{gray}{\(\pm 0.008\)}}} &
\shortstack{\(\mathbf{0.904}\)\\{\tiny\textcolor{gray}{\(\pm 0.006\)}}} &
\shortstack{\(0.716\)\\{\tiny\textcolor{gray}{\(\pm 0.005\)}}} &
\shortstack{\(\mathbf{0.138}\)\\{\tiny\textcolor{gray}{\(\pm 0.048\)}}} &
\shortstack{\(\mathbf{0.739}\)\\{\tiny\textcolor{gray}{\(\pm 0.067\)}}} &
\shortstack{\(\mathbf{0.476}\)\\{\tiny\textcolor{gray}{\(\pm 0.090\)}}} &
\shortstack{\(\mathbf{0.467}\)\\{\tiny\textcolor{gray}{\(\pm 0.011\)}}} &
\shortstack{\(\mathbf{2.956}\)\\{\tiny\textcolor{gray}{\(\pm 0.080\)}}} &
\shortstack{\(790.9\)\\{\tiny\textcolor{gray}{\(\pm 18.53\)}}} & 
\shortstack{\(\mathbf{7.28}\)\\{\tiny\textcolor{gray}{\(\pm 0.15\)}}} &
\shortstack{\(4439.8\)\\{\tiny\textcolor{gray}{\(\pm 71.2\)}}} &
\shortstack{\(\mathbf{1.018}\)\\{\tiny\textcolor{gray}{\(\pm 0.018\)}}} \\

\midrule
\multicolumn{18}{l}{\textit{\textbf{Other methods}}} \\
\makecell[l]{Ours [TabPFN+CART]\\[2pt]}     & 
\shortstack{\(\mathbf{0.828}\)\\{\tiny\textcolor{gray}{\(\pm 0.008\)}}} &
\shortstack{\(0.917\)\\{\tiny\textcolor{gray}{\(\pm 0.009\)}}} &
\shortstack{\(\mathbf{0.792}\)\\{\tiny\textcolor{gray}{\(\pm 0.006\)}}} &
\shortstack{\(0.843\)\\{\tiny\textcolor{gray}{\(\pm 0.010\)}}} &
\shortstack{\(\mathbf{0.738}\)\\{\tiny\textcolor{gray}{\(\pm 0.007\)}}} &
\shortstack{\(\mathbf{0.937}\)\\{\tiny\textcolor{gray}{\(\pm 0.005\)}}} &
\shortstack{\(\mathbf{0.990}\)\\{\tiny\textcolor{gray}{\(\pm 0.001\)}}} &
\shortstack{\(0.740\)\\{\tiny\textcolor{gray}{\(\pm 0.010\)}}} &
\shortstack{\(\mathbf{0.014}\)\\{\tiny\textcolor{gray}{\(\pm 0.002\)}}} &
\shortstack{\(0.426\)\\{\tiny\textcolor{gray}{\(\pm 0.025\)}}} &
\shortstack{\(\mathbf{0.218}\)\\{\tiny\textcolor{gray}{\(\pm 0.013\)}}} &
\shortstack{\(0.445\)\\{\tiny\textcolor{gray}{\(\pm 0.010\)}}} &
\shortstack{\(\mathbf{1.029}\)\\{\tiny\textcolor{gray}{\(\pm 0.041\)}}} &
\shortstack{\(722.8\)\\{\tiny\textcolor{gray}{\(\pm 9.79\)}}} &
\shortstack{\(4.03\)\\{\tiny\textcolor{gray}{\(\pm 0.03\)}}} &
\shortstack{\(4620.0\)\\{\tiny\textcolor{gray}{\(\pm 266.9\)}}} &
\shortstack{\(\mathbf{0.849}\)\\{\tiny\textcolor{gray}{\(\pm 0.026\)}}} \\

\makecell[l]{DeLTA [RF+CART]\\[2pt]} & 
\shortstack{\(0.738\)\\{\tiny\textcolor{gray}{\(\pm 0.010\)}}} & 
\shortstack{\(0.846\)\\{\tiny\textcolor{gray}{\(\pm 0.012\)}}} & 
\shortstack{\(0.742\)\\{\tiny\textcolor{gray}{\(\pm 0.009\)}}} & 
\shortstack{\(0.774\)\\{\tiny\textcolor{gray}{\(\pm 0.011\)}}} & 
\shortstack{\(0.649\)\\{\tiny\textcolor{gray}{\(\pm 0.014\)}}} & 
\shortstack{\(0.906\)\\{\tiny\textcolor{gray}{\(\pm 0.005\)}}} & 
\shortstack{\(0.947\)\\{\tiny\textcolor{gray}{\(\pm 0.004\)}}} & 
\shortstack{\(0.613\)\\{\tiny\textcolor{gray}{\(\pm 0.008\)}}} & 
\shortstack{\(0.915\)\\{\tiny\textcolor{gray}{\(\pm 0.002\)}}} & 
\shortstack{\(0.742\)\\{\tiny\textcolor{gray}{\(\pm 0.010\)}}} & 
\shortstack{\(0.647\)\\{\tiny\textcolor{gray}{\(\pm 0.004\)}}} & 
\shortstack{\(1.438\)\\{\tiny\textcolor{gray}{\(\pm 0.003\)}}} & 
\shortstack{\(2.403\)\\{\tiny\textcolor{gray}{\(\pm 0.019\)}}} & 
\shortstack{\(779.2\)\\{\tiny\textcolor{gray}{\(\pm 11.09\)}}} & 
\shortstack{\(5.98\)\\{\tiny\textcolor{gray}{\(\pm 0.13\)}}} & 
\shortstack{\(5165.4\)\\{\tiny\textcolor{gray}{\(\pm 107.9\)}}} & 
\shortstack{\(0.985\)\\{\tiny\textcolor{gray}{\(\pm 0.028\)}}} \\

\makecell[l]{DeLTA [RF+TabPFN]\\[2pt]} & 
\shortstack{\(0.797\)\\{\tiny\textcolor{gray}{\(\pm 0.008\)}}} & 
\shortstack{\(0.883\)\\{\tiny\textcolor{gray}{\(\pm 0.006\)}}} & 
\shortstack{\(0.774\)\\{\tiny\textcolor{gray}{\(\pm 0.008\)}}} & 
\shortstack{\(0.809\)\\{\tiny\textcolor{gray}{\(\pm 0.008\)}}} & 
\shortstack{\(0.706\)\\{\tiny\textcolor{gray}{\(\pm 0.012\)}}} & 
\shortstack{\(0.923\)\\{\tiny\textcolor{gray}{\(\pm 0.006\)}}} & 
\shortstack{\(0.974\)\\{\tiny\textcolor{gray}{\(\pm 0.003\)}}} & 
\shortstack{\(0.658\)\\{\tiny\textcolor{gray}{\(\pm 0.007\)}}} & 
\shortstack{\(0.921\)\\{\tiny\textcolor{gray}{\(\pm 0.002\)}}} & 
\shortstack{\(0.738\)\\{\tiny\textcolor{gray}{\(\pm 0.006\)}}} & 
\shortstack{\(0.628\)\\{\tiny\textcolor{gray}{\(\pm 0.003\)}}} & 
\shortstack{\(1.432\)\\{\tiny\textcolor{gray}{\(\pm 0.004\)}}} & 
\shortstack{\(1.357\)\\{\tiny\textcolor{gray}{\(\pm 0.042\)}}} & 
\shortstack{\(\mathbf{719.7}\)\\{\tiny\textcolor{gray}{\(\pm 9.93\)}}} & 
\shortstack{\(4.82\)\\{\tiny\textcolor{gray}{\(\pm 0.13\)}}} & 
\shortstack{\(\mathbf{4542.5}\)\\{\tiny\textcolor{gray}{\(\pm 88.7\)}}} &
\shortstack{\(0.921\)\\{\tiny\textcolor{gray}{\(\pm 0.030\)}}} \\

\makecell[l]{OCTree [XGBoost]\\[2pt]}       & 
\shortstack{\(0.815\)\\{\tiny\textcolor{gray}{\(\pm 0.012\)}}} &
\shortstack{\(0.872\)\\{\tiny\textcolor{gray}{\(\pm 0.004\)}}} &
\shortstack{\(0.775\)\\{\tiny\textcolor{gray}{\(\pm 0.008\)}}} &
\shortstack{\(0.823\)\\{\tiny\textcolor{gray}{\(\pm 0.004\)}}} &
\shortstack{\(0.705\)\\{\tiny\textcolor{gray}{\(\pm 0.013\)}}} &
\shortstack{\(0.918\)\\{\tiny\textcolor{gray}{\(\pm 0.005\)}}} &
\shortstack{\(0.959\)\\{\tiny\textcolor{gray}{\(\pm 0.001\)}}} &
\shortstack{\(0.692\)\\{\tiny\textcolor{gray}{\(\pm 0.007\)}}} &
\shortstack{\(0.037\)\\{\tiny\textcolor{gray}{\(\pm 0.013\)}}} &
\shortstack{\(0.587\)\\{\tiny\textcolor{gray}{\(\pm 0.030\)}}} &
\shortstack{\(0.285\)\\{\tiny\textcolor{gray}{\(\pm 0.021\)}}} &
\shortstack{\(0.468\)\\{\tiny\textcolor{gray}{\(\pm 0.017\)}}} &
\shortstack{\(2.514\)\\{\tiny\textcolor{gray}{\(\pm 0.099\)}}} &
\shortstack{\(764.3\)\\{\tiny\textcolor{gray}{\(\pm 11.32\)}}} &
\shortstack{\(4.87\)\\{\tiny\textcolor{gray}{\(\pm 0.19\)}}} &
\shortstack{\(6380.8\)\\{\tiny\textcolor{gray}{\(\pm 1040.9\)}}} &
\shortstack{\(0.879\)\\{\tiny\textcolor{gray}{\(\pm 0.008\)}}}
 \\

\makecell[l]{XGBoost [TabArena]\\[2pt]}   & 
\shortstack{\(0.814\)\\{\tiny\textcolor{gray}{\(\pm 0.007\)}}} &
\shortstack{\(0.907\)\\{\tiny\textcolor{gray}{\(\pm 0.008\)}}} &
\shortstack{\(0.788\)\\{\tiny\textcolor{gray}{\(\pm 0.005\)}}} &
\shortstack{\(0.839\)\\{\tiny\textcolor{gray}{\(\pm 0.007\)}}} &
\shortstack{\(\mathbf{0.738}\)\\{\tiny\textcolor{gray}{\(\pm 0.008\)}}} &
\shortstack{\(0.934\)\\{\tiny\textcolor{gray}{\(\pm 0.008\)}}} &
\shortstack{\(0.975\)\\{\tiny\textcolor{gray}{\(\pm 0.002\)}}} &
\shortstack{\(0.719\)\\{\tiny\textcolor{gray}{\(\pm 0.016\)}}} &
\shortstack{\(0.023\)\\{\tiny\textcolor{gray}{\(\pm 0.008\)}}} &
\shortstack{\(0.423\)\\{\tiny\textcolor{gray}{\(\pm 0.018\)}}} &
\shortstack{\(0.255\)\\{\tiny\textcolor{gray}{\(\pm 0.007\)}}} &
\shortstack{\(0.446\)\\{\tiny\textcolor{gray}{\(\pm 0.008\)}}} &
\shortstack{\(1.406\)\\{\tiny\textcolor{gray}{\(\pm 0.041\)}}} &
\shortstack{\(745.1\)\\{\tiny\textcolor{gray}{\(\pm 10.57\)}}} &
\shortstack{\(4.17\)\\{\tiny\textcolor{gray}{\(\pm 0.04\)}}} &
\shortstack{\(4580.4\)\\{\tiny\textcolor{gray}{\(\pm 231.5\)}}} &
\shortstack{\(0.870\)\\{\tiny\textcolor{gray}{\(\pm 0.023\)}}} \\

\makecell[l]{TabM [TabArena]\\[2pt]}      & 
\shortstack{\(0.822\)\\{\tiny\textcolor{gray}{\(\pm 0.006\)}}} &
\shortstack{\(0.917\)\\{\tiny\textcolor{gray}{\(\pm 0.006\)}}} &
\shortstack{\(\mathbf{0.792}\)\\{\tiny\textcolor{gray}{\(\pm 0.009\)}}} &
\shortstack{\(0.838\)\\{\tiny\textcolor{gray}{\(\pm 0.007\)}}} &
\shortstack{\(\mathbf{0.738}\)\\{\tiny\textcolor{gray}{\(\pm 0.004\)}}} &
\shortstack{\(0.935\)\\{\tiny\textcolor{gray}{\(\pm 0.008\)}}} &
\shortstack{\(0.982\)\\{\tiny\textcolor{gray}{\(\pm 0.002\)}}} &
\shortstack{\(0.723\)\\{\tiny\textcolor{gray}{\(\pm 0.014\)}}} &
\shortstack{\(0.020\)\\{\tiny\textcolor{gray}{\(\pm 0.007\)}}} &
\shortstack{\(0.450\)\\{\tiny\textcolor{gray}{\(\pm 0.020\)}}} &
\shortstack{\(0.240\)\\{\tiny\textcolor{gray}{\(\pm 0.015\)}}} &
\shortstack{\(\mathbf{0.438}\)\\{\tiny\textcolor{gray}{\(\pm 0.007\)}}} &
\shortstack{\(1.114\)\\{\tiny\textcolor{gray}{\(\pm 0.042\)}}} &
\shortstack{\(753.6\)\\{\tiny\textcolor{gray}{\(\pm 10.06\)}}} &
\shortstack{\(4.25\)\\{\tiny\textcolor{gray}{\(\pm 0.16\)}}} &
\shortstack{\(4585.6\)\\{\tiny\textcolor{gray}{\(\pm 211.2\)}}} &
\shortstack{\(0.886\)\\{\tiny\textcolor{gray}{\(\pm 0.017\)}}} \\

\makecell[l]{TabPFNv2 [TabArena]\\[2pt]}            & 
\shortstack{\(0.824\)\\{\tiny\textcolor{gray}{\(\pm 0.005\)}}} &
\shortstack{\(\mathbf{0.919}\)\\{\tiny\textcolor{gray}{\(\pm 0.007\)}}} &
\shortstack{\(0.761\)\\{\tiny\textcolor{gray}{\(\pm 0.006\)}}} &
\shortstack{\(\mathbf{0.846}\)\\{\tiny\textcolor{gray}{\(\pm 0.003\)}}} &
\shortstack{\(\mathbf{0.738}\)\\{\tiny\textcolor{gray}{\(\pm 0.006\)}}} &
\shortstack{\(0.934\)\\{\tiny\textcolor{gray}{\(\pm 0.008\)}}} &
\shortstack{\(0.988\)\\{\tiny\textcolor{gray}{\(\pm 0.001\)}}} &
\shortstack{\(\mathbf{0.742}\)\\{\tiny\textcolor{gray}{\(\pm 0.015\)}}} &
\shortstack{\(0.021\)\\{\tiny\textcolor{gray}{\(\pm 0.004\)}}} &
\shortstack{\(\mathbf{0.413}\)\\{\tiny\textcolor{gray}{\(\pm 0.019\)}}} &
\shortstack{\(0.223\)\\{\tiny\textcolor{gray}{\(\pm 0.011\)}}} &
\shortstack{\(0.445\)\\{\tiny\textcolor{gray}{\(\pm 0.011\)}}} &
\shortstack{\(1.045\)\\{\tiny\textcolor{gray}{\(\pm 0.033\)}}} &
\shortstack{\(736.1\)\\{\tiny\textcolor{gray}{\(\pm 14.18\)}}} &
\shortstack{\(\mathbf{4.00}\)\\{\tiny\textcolor{gray}{\(\pm 0.03\)}}} &
\shortstack{\(4717.3\)\\{\tiny\textcolor{gray}{\(\pm 247.3\)}}} &
\shortstack{\(0.854\)\\{\tiny\textcolor{gray}{\(\pm 0.028\)}}} \\

\bottomrule
\end{tabular}
\end{adjustbox}
\vspace{-20px}
\end{table*}

\vspace{-10pt}
\section{Performance Evaluation}\label{sect:experiments}
\vspace{-8pt}
\subsection{Results and Discussion}\label{sect:exp_main_results}
\vspace{-5pt}

\textbf{Setup.}
We evaluate on low-resource tabular problems from OpenML~\cite{vanschoren2013openml,feurer2019openmlpython,bischl2025openml}. Specifically, we use all 17 supervised datasets from \texttt{tabarena-v0.1}~\cite{erickson2025tabarena} containing at most 2500 samples: 8 binary classification, 4 multiclass classification, and 5 regression. We report ROC AUC$\uparrow$ for binary classification, LogLoss$\downarrow$ for multiclass, and RMSE$\downarrow$ for regression. For each dataset, we use the first 5 canonical OpenML splits (\texttt{repeat=0\dots4}) with a stratified 80\%/20\% train/validation split, reporting both the average score and the standard error of the mean over 5 runs in the main table. Additional train, validation, test, hard-label classification, depth, leaves, and runtime statistics are reported in Appendix~\ref{app:optimal_tree_results}. Full dataset information appears in Appendix~\ref{app:datasets_info}.

Our method uses GPT-5 (default thinking budget) for agentic tree construction. The LLM has access to train--validation splits but not the test set. We compare against several baselines. For traditional tabular methods, we include: CART decision trees from \texttt{Scikit-Learn} (both default and tuned), TAO~\cite{carreira2018alternating} as a recent non-greedy tree learner, XGBoost~\cite{chen2016xgboost} (tuned GBDT), TabM~\cite{gorishniy2025_tabm} (advanced NN ensemble), and TabPFN v2~\cite{hollmann2022tabpfn,hoo2025_tabpfn_ts} (tabular foundation model). CART (tuned) optimizes validation score over maximum depth, minimum samples per split and leaf, and splitting criterion. For LLM-based methods, we include OCTree~\cite{nam2024optimized}, which performs LLM-driven feature generation followed by XGBoost training, and DeLTA~\cite{ye2025_llm_meeting_decision_trees}, which uses an LLM to edit random-forest trees and fits either CART or TabPFN to each partition. Hyperparameters for all methods appear in Appendix~\ref{app:model_hyperparams}. For broader coverage of the optimal-tree literature, we additionally evaluate SAT-based tree improvement~\cite{narodytska2018sat} and DPDT~\cite{kohler2025breiman}; both return only hard labels for classification (and SAT does not support regression), so we report their full results under the appropriate metrics in Appendix~\ref{app:optimal_tree_results}.

We also evaluate a variant where agentic decision trees correct TabPFN v2 predictions. Here, the LLM builds a regression tree on out-of-fold TabPFN v2 predictions from training and validation data; tree outputs are added to TabPFN v2 test predictions. For classification, we normalize corrected predictions to sum to 1 for LogLoss evaluation.

\textbf{Comparison to Tabular Methods.}
Table~\ref{tab:main_table} (top panel) shows that LLM-made decision trees surpass both default and tuned CART on all but one dataset. The exception is UsedFiat500, where our model underperforms the tuned baseline by 0.1\%. The agentic tree also remains competitive with TAO under the primary probabilistic and regression metrics, while improving over it on a majority of datasets. Additional appendix results show that TAO, DPDT, and SAT improve over vanilla CART on several datasets under hard-label metrics, but the agentic tree retains strong test performance with a single-tree representation. Table~\ref{tab:main_table} (bottom panel) shows that black-box methods (XGBoost, TabM, TabPFN v2) still outperform individual decision trees on most problems, but agentic tree learning significantly reduces this gap. Using agentic decision trees to correct TabPFN v2 predictions improves performance in most cases. While this two-stage pipeline is not interpretable (due to TabPFN), it proves useful when the agentic tree can correct biases or incorporate additional inputs such as feature descriptions or domain knowledge (see Section~\ref{sec:control}).

\textbf{Comparison to LLM-Based Methods.}
OCTree and DeLTA differ significantly in how they employ LLMs: OCTree iteratively proposes and implements new features, while DeLTA partitions the data space. Crucially, OCTree uses XGBoost as its final decision layer, and DeLTA fits a random forest followed by either CART or TabPFN. Our agentic formulation generalizes OCTree's feature engineering approach, since the LLM can and does perform feature engineering (Figure~\ref{fig:func-call-distr}). Because OCTree uses XGBoost, the most appropriate comparison is with our TabPFN-based version, which outperforms OCTree on all datasets. Notably, OCTree decreases performance relative to standalone XGBoost on almost all datasets. We attribute this to two factors: first, OCTree must repeatedly search over candidate features while the underlying XGBoost cannot be fully re-tuned (e.g., via Optuna) at every iteration, creating a mismatch between feature search and final configuration; second, since most datasets contain fewer than 2,500 samples, introducing additional features tends to increase the overfitting effect.
For DeLTA, there are two comparison modes: CART and TabPFN versions. Our single-tree implementation outperforms DeLTA with CART on 9 of 17 datasets, and our TabPFN version outperforms DeLTA with TabPFN on all but two datasets.

\textbf{Runtime and complexity.}
Runtime exposes an important interpretability--performance tradeoff. CART is much faster than all iterative methods, while TAO and DPDT typically run close to the 300s tuning budget and SAT is slower on classification tasks and unavailable for regression. DPDT tends to produce the most compact trees, TAO often yields larger trees, and the agentic trees occupy a compact middle range while achieving strong test performance. Full train, validation, test, hard-label classification, depth, leaves, and runtime statistics are given in Appendix~\ref{app:optimal_tree_results}.

\vspace{-8pt}
\subsection{Ablation}\label{sect:exp_ablation}
\begin{table*}[t]
  \centering
  \scriptsize
  \setlength{\tabcolsep}{2pt}
  \renewcommand{\arraystretch}{1.05}
  \caption{\textbf{Left:} Model Comparison, \textbf{Right:} Ablation Study.}
  \label{tab:ablations}

  \begin{minipage}[t]{0.49\linewidth}
    \centering
    \begin{tabular}{lccc}
      \toprule
      & \makecell{Fitness\\[-0.3ex]{\tiny\textcolor{gray}{ROC AUC$\uparrow$}}}
      & \makecell{Anneal\\[-0.3ex]{\tiny\textcolor{gray}{LogLoss$\downarrow$}}}
      & \makecell{Airfoil\\[-0.3ex]{\tiny\textcolor{gray}{RMSE$\downarrow$}}} \\
      \midrule
      Main [GPT‑5]           &
        \shortstack{\(0.818\)\\[-0.4ex]{\tiny\textcolor{gray}{\(\pm 0.017\)}}} &
        \shortstack{\(0.138\)\\[-0.4ex]{\tiny\textcolor{gray}{\(\pm 0.108\)}}} &
        \shortstack{\(2.956\)\\[-0.4ex]{\tiny\textcolor{gray}{\(\pm 0.180\)}}} \\
      Gemini 2.5 Pro         &
        \shortstack{\(0.800\)\\[-0.4ex]{\tiny\textcolor{gray}{\(\pm 0.011\)}}} &
        \shortstack{\(0.624\)\\[-0.4ex]{\tiny\textcolor{gray}{\(\pm 0.552\)}}} &
        \shortstack{\(4.004\)\\[-0.4ex]{\tiny\textcolor{gray}{\(\pm 0.168\)}}} \\
      Claude Opus 4.5         &
        \shortstack{\(0.816\)\\[-0.4ex]{\tiny\textcolor{gray}{\(\pm 0.015\)}}} &
        \shortstack{\(0.184\)\\[-0.4ex]{\tiny\textcolor{gray}{\(\pm 0.135\)}}} &
        \shortstack{\(2.982\)\\[-0.4ex]{\tiny\textcolor{gray}{\(\pm 0.157\)}}} \\
      GPT-5.2                &
        \shortstack{\(0.812\)\\[-0.4ex]{\tiny\textcolor{gray}{\(\pm 0.013\)}}} &
        \shortstack{\(0.210\)\\[-0.4ex]{\tiny\textcolor{gray}{\(\pm 0.248\)}}} &
        \shortstack{\(3.940\)\\[-0.4ex]{\tiny\textcolor{gray}{\(\pm 0.530\)}}} \\
      GLM 4.5                &
        \shortstack{\(0.799\)\\[-0.4ex]{\tiny\textcolor{gray}{\(\pm 0.015\)}}} &
        \shortstack{\(0.224\)\\[-0.4ex]{\tiny\textcolor{gray}{\(\pm 0.113\)}}} &
        \shortstack{\(3.620\)\\[-0.4ex]{\tiny\textcolor{gray}{\(\pm 0.394\)}}} \\
      Kimi K2 Instruct       &
        \shortstack{\(0.808\)\\[-0.4ex]{\tiny\textcolor{gray}{\(\pm 0.014\)}}} &
        \shortstack{\(0.161\)\\[-0.4ex]{\tiny\textcolor{gray}{\(\pm 0.056\)}}} &
        \shortstack{\(3.764\)\\[-0.4ex]{\tiny\textcolor{gray}{\(\pm 0.505\)}}} \\
      DeepSeek R1 0528        &
        \shortstack{\(0.797\)\\[-0.4ex]{\tiny\textcolor{gray}{\(\pm 0.011\)}}} &
        \shortstack{\(0.410\)\\[-0.4ex]{\tiny\textcolor{gray}{\(\pm 0.320\)}}} &
        \shortstack{\(3.874\)\\[-0.4ex]{\tiny\textcolor{gray}{\(\pm 0.733\)}}} \\
      \bottomrule
    \end{tabular}
  \end{minipage}
  \hfill
  \begin{minipage}[t]{0.49\linewidth}
    \centering
    \begin{tabular}{lccc}
      \toprule
      & \makecell{Fitness\\[-0.3ex]{\tiny\textcolor{gray}{ROC AUC$\uparrow$}}}
      & \makecell{Anneal\\[-0.3ex]{\tiny\textcolor{gray}{LogLoss$\downarrow$}}}
      & \makecell{Airfoil\\[-0.3ex]{\tiny\textcolor{gray}{RMSE$\downarrow$}}} \\
      \midrule
      Main method            &
        \shortstack{\(0.818\)\\[-0.4ex]{\tiny\textcolor{gray}{\(\pm 0.017\)}}} &
        \shortstack{\(0.138\)\\[-0.4ex]{\tiny\textcolor{gray}{\(\pm 0.108\)}}} &
        \shortstack{\(2.956\)\\[-0.4ex]{\tiny\textcolor{gray}{\(\pm 0.180\)}}} \\
      CART+AUG               &
        \shortstack{\(0.791\)\\[-0.4ex]{\tiny\textcolor{gray}{\(\pm 0.034\)}}} &
        \shortstack{\(0.246\)\\[-0.4ex]{\tiny\textcolor{gray}{\(\pm 0.008\)}}} &
        \shortstack{\(5.542\)\\[-0.4ex]{\tiny\textcolor{gray}{\(\pm 0.079\)}}} \\
      CART+AG                &
        \shortstack{\(0.802\)\\[-0.4ex]{\tiny\textcolor{gray}{\(\pm 0.027\)}}} &
        \shortstack{\(0.174\)\\[-0.4ex]{\tiny\textcolor{gray}{\(\pm 0.000\)}}} &
        \shortstack{\(2.814\)\\[-0.4ex]{\tiny\textcolor{gray}{\(\pm 0.219\)}}} \\
      Unlimited (not tree‑only) &
        \shortstack{\(0.807\)\\[-0.4ex]{\tiny\textcolor{gray}{\(\pm 0.006\)}}} &
        \shortstack{\(0.060\)\\[-0.4ex]{\tiny\textcolor{gray}{\(\pm 0.056\)}}} &
        \shortstack{\(1.621\)\\[-0.4ex]{\tiny\textcolor{gray}{\(\pm 0.094\)}}} \\
      Agent‑sklearn trees    &
        \shortstack{\(0.795\)\\[-0.4ex]{\tiny\textcolor{gray}{\(\pm 0.014\)}}} &
        \shortstack{\(0.257\)\\[-0.4ex]{\tiny\textcolor{gray}{\(\pm 0.230\)}}} &
        \shortstack{\(2.943\)\\[-0.4ex]{\tiny\textcolor{gray}{\(\pm 0.705\)}}} \\
      Main w/o description   &
        \shortstack{\(0.813\)\\[-0.4ex]{\tiny\textcolor{gray}{\(\pm 0.011\)}}} &
        \shortstack{\(0.125\)\\[-0.4ex]{\tiny\textcolor{gray}{\(\pm 0.043\)}}} &
        \shortstack{\(3.508\)\\[-0.4ex]{\tiny\textcolor{gray}{\(\pm 0.438\)}}} \\
      Data leak             &
        \shortstack{\(0.515\)\\[-0.4ex]{\tiny\textcolor{gray}{\(\pm 0.050\)}}} &
        \shortstack{\(0.845\)\\[-0.4ex]{\tiny\textcolor{gray}{\(\pm 0.013\)}}} &
        \shortstack{\(7.216\)\\[-0.4ex]{\tiny\textcolor{gray}{\(\pm 0.228\)}}} \\
      No agent              &
        \shortstack{\(0.692\)\\[-0.4ex]{\tiny\textcolor{gray}{\(\pm 0.012\)}}} &
        \shortstack{\(0.541\)\\[-0.4ex]{\tiny\textcolor{gray}{\(\pm 0.005\)}}} &
        \shortstack{\(5.355\)\\[-0.4ex]{\tiny\textcolor{gray}{\(\pm 0.132\)}}} \\
      \bottomrule
    \end{tabular}
  \end{minipage}
  \vspace{-10pt}
\end{table*}
\vspace{-5pt}
\textbf{CART+AUG and CART+AG.} To control for non-agentic alternatives, we compare against two non-agentic CART variants: \textbf{CART+AUG}, which tunes CART together with generic feature augmentations (additive and multiplicative interactions, sparse feature selection, PCA), and \textbf{CART+AG}, which trains a tuned CART on the representation produced by the agent, isolating how much of the gain comes from preprocessing rather than from the final tree edits. Both controls often improve over the no-agent baseline and can be competitive on individual datasets, but they do not explain the overall gains: the agentic tree remains best on Fitness and Anneal in this slice (Table~\ref{tab:ablations}), and the full 17-dataset comparison in Appendix~\ref{app:optimal_tree_results} shows it outperforms CART+AUG and CART+AG on most primary metrics.

\textbf{Unrestricted.} We grant the LLM unrestricted access to the entire \texttt{Scikit-Learn} library, challenging it to construct advanced model ensembles not limited to decision trees. As such, the LLM can utilize shallow MLPs, kernel machines, splines, and generalized additive models. Unsurprisingly, the unrestricted LLM agent can outperform individual trees on harder datasets (Anneal, Airfoil) by using ensembles. However, it still fails to compete with advanced black-box methods in Table~\ref{tab:main_table}.

\textbf{No tree scaffold.} We replicate the tree-building experiments from Table~\ref{tab:main_table} but using only the \texttt{Scikit-Learn} tree frameworks. From Table~\ref{tab:ablations}, the tree agent without our editing tools consistently underperforms across 3 datasets, exhibiting high standard deviation on Airfoil and Anneal. Examining thought traces, we find that the agent has trouble manipulating the tree and instead falls back to data preprocessing.

\textbf{No metadata.} We evaluate the importance of supplementary dataset metadata (e.g., description and feature names) provided to the LLM. As reported in Table~\ref{tab:ablations}, giving the LLM access to informal descriptions generally tends to improve performance. However, in a separate experiment reported in Appendix~\ref{app:data_leak}, providing intentionally incorrect or mismatched descriptions can mislead the model and hurt performance.

\textbf{Data leaks.} We define cheating as a \emph{recognition-and-retrieval} behavior, where the LLM recognizes a benchmark and edits the tree to match a memorized solution or memorized test set, instead of fitting to the provided training labels. To check for this, we keep the true dataset description and feature marginals intact but erase the train/validation dependence structure by shuffling values within each column and permuting training labels. This leaves summaries and small printed snippets unchanged, so dataset recognition triggers, while genuine learning from the training set becomes impossible. Under this trap condition, the agent's clean-test performance drops to near-baseline, which is inconsistent with memorization-driven shortcuts. A more comprehensive ablation over metadata and corruption levels, together with additional baselines, is reported in Appendix~\ref{app:data_leak}.

\textbf{No agent setup.} To quantify how much our gains rely on \emph{agentic} interaction, we remove the iterative agent loop and instead ask the LLM to produce an ``optimal'' edited decision tree for a given dataset in a single zero-shot response (i.e., no hypothesis--test cycle, no intermediate diagnostics, and no multi-step refinement). As shown in Table~\ref{tab:ablations}, the absence of the agent loop substantially degrades performance across all three tasks. This gap indicates that the agentic scaffold is a key driver of the final performance, and that zero-shot ``one-pass'' tree editing is insufficient.

\textbf{LLM Backbone.} We perform a detailed comparison of different backbone LLMs. For clarity, we only vary the base LLM used within our agent while keeping the agentic framework, toolset, and prompts fixed. We evaluate four closed-source models (GPT-5, GPT-5.2, Gemini 2.5 Pro, Claude Opus 4.5) and three open-source models (GLM-4.5, Kimi K2 Instruct, DeepSeek R1-0528). On more challenging or low-signal tasks (Anneal and Airfoil), we observe the greatest variability, with GPT-5 achieving the strongest overall performance. Among the closed-source competitors, Claude Opus 4.5 is the closest. Among the open-source models, Kimi K2 Instruct takes the lead. As the original prompt was developed for GPT-5, these results should be interpreted as a lower bound for the other backbones under an untuned prompt configuration.
\vspace{-8pt}
\subsection{Agent operations}\label{sect:exp_analysis}
\vspace{-5pt}
\begin{figure*}[t]
    \vspace{-5px}
    \centering
    \includegraphics[width=0.85\linewidth]{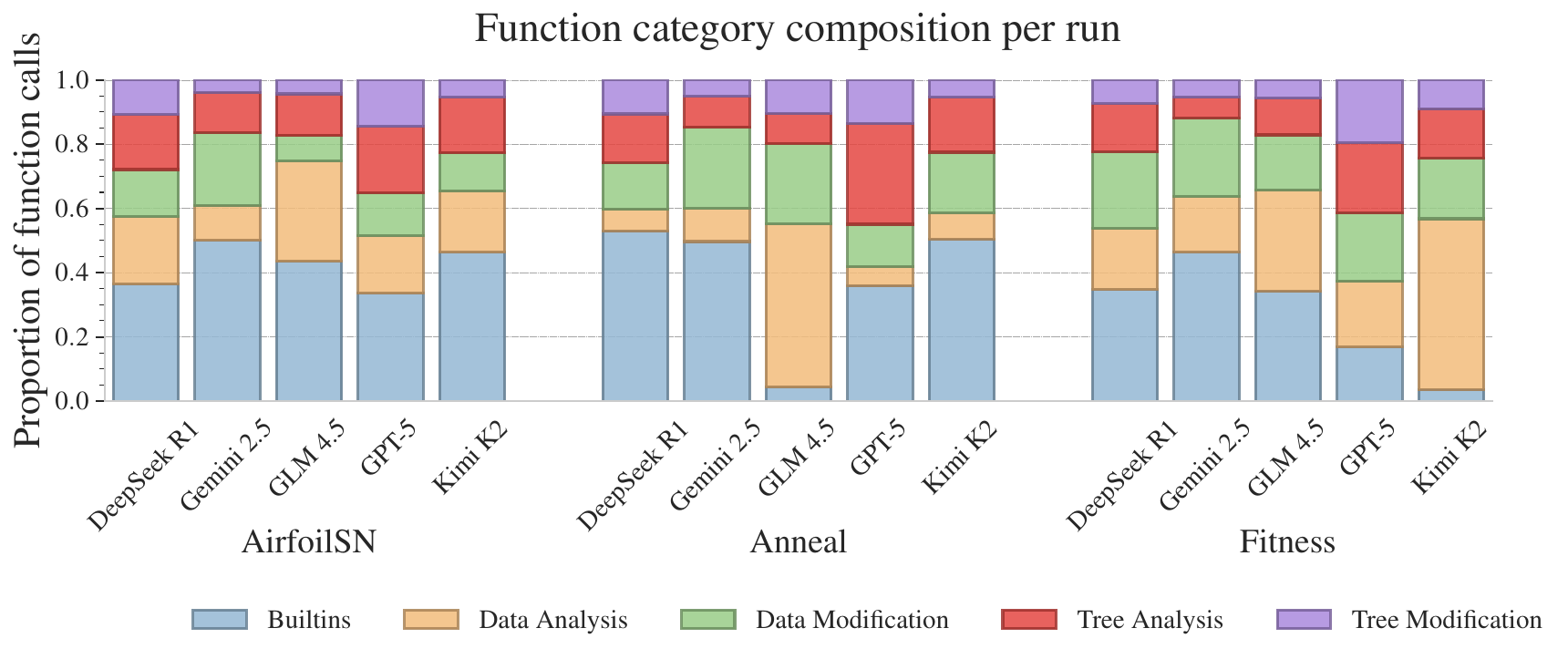}
    \vspace{-5px}
    \caption{Tool call distribution across LLM backbones and datasets categorized by functionality.}
    \label{fig:func-call-distr}
    \vspace{-10px}
\end{figure*}

In this section, we analyze the tree construction process in more detail to better understand current agent behavior and inform potential future improvements. We analyze the code generated during each ``action'' phase, grouping function calls into categories. We categorize the agent's actions into the following five groups by their operational target (the model vs.\ the data) and their purpose (modification vs.\ analysis): (1) \textbf{tree modification} for training or altering the model; (2) \textbf{tree analysis} for model introspection; (3) \textbf{data analysis} for dataset-level exploration; (4) \textbf{data modification} for data cleaning, feature engineering, and similar; and (5) \textbf{builtins} for supporting functions.

We illustrate our findings in Figure~\ref{fig:func-call-distr}. The distribution reveals two broad regularities: (1) exploratory analysis and feature engineering together account for roughly half of all calls, and (2) generic helper/builtins contribute a comparable share, while direct tree edits (grafting/pruning/growing) remain a small tail, typically under 10\%. In other words, the LLM agents typically perform a lot of analysis and testing before every edit. We also observe substantial model- and dataset-specific variability. GPT-5 exhibits the most balanced mix, with no single family dominating across tasks. By contrast, Gemini 2.5 Pro --- the weakest model in this specific setting --- leans heavily on feature engineering and helper utilities, and consistently deprioritizes tree construction, with only about 5\% of its calls devoted to explicit tree modifications. GLM-4.5, Kimi K2 Instruct, and DeepSeek R1-0528 similarly skew toward general data-analysis operations. For context, we also provide word clouds for each function category in Appendix~\ref{app:agent-analysis}.

\vspace{-8pt}
\section{Controllability Analysis}\label{sec:control}
\vspace{-8pt}
One of the unique advantages of LLM-based tree construction is that the task can be defined informally, through the prompt. This allows the user to specify additional requirements that can be difficult or impossible to formalize for traditional decision trees and even for more advanced tabular methods. In this section, we consider three such scenarios: (1) \textbf{informal feature definition}, where the system was just given access to a new feature that is not yet present in historical data; (2) \textbf{fairness:} correcting bias in the data; and (3) \textbf{monotonic constraints:} making the decision function monotonic with respect to a set of features.

\vspace{-8pt}
\subsection{Missing Feature}\label{subsec:lost_feature}
\vspace{-5pt}
First, we consider a less studied setup that, nonetheless, often happens in real-world applications. In this setup, a machine learning model has access to some feature during deployment but does not have this feature in the training data. This can occur in several ways: (1) cold-start situations where a new feature is added; (2) legal feature-storage bans that prevent storing the information. In both cases, some important features are missing in the training data, but there is still informal domain knowledge about how that feature can help prediction. We feed this knowledge as part of the LLM agent's prompt and ask it to incorporate the feature into the tree without data. Then, we evaluate the final tree on a dataset with the extra feature. For this evaluation, we take the \texttt{Diabetes} dataset and hide the \texttt{Glucose} feature from the agent's training data. Instead, we provide only an informal note (see Appendix~\ref{app:prompts_human_input}) on the existence of the missing feature, while any understanding of its relevance to the prediction target must come from the model's prior knowledge.
We compare four setups:
\begin{itemize}[leftmargin=*]
    \vspace{-4px}\item \textbf{Ours: feature in data} -- LLM-built tree trained on the dataset with the feature present \textit{in the training data};
    \vspace{-4px}\item \textbf{Ours: feature in prompt} -- LLM-built tree trained with the \emph{feature in the prompt};
    \vspace{-4px}\item \textbf{Ours: no feature} -- no feature in prompt or data;
    \vspace{-4px}\item \textbf{CART sklearn} decision tree without the feature.
\end{itemize}
\vspace{-4px}

\begin{figure*}[t]
  \includegraphics[width=\textwidth]{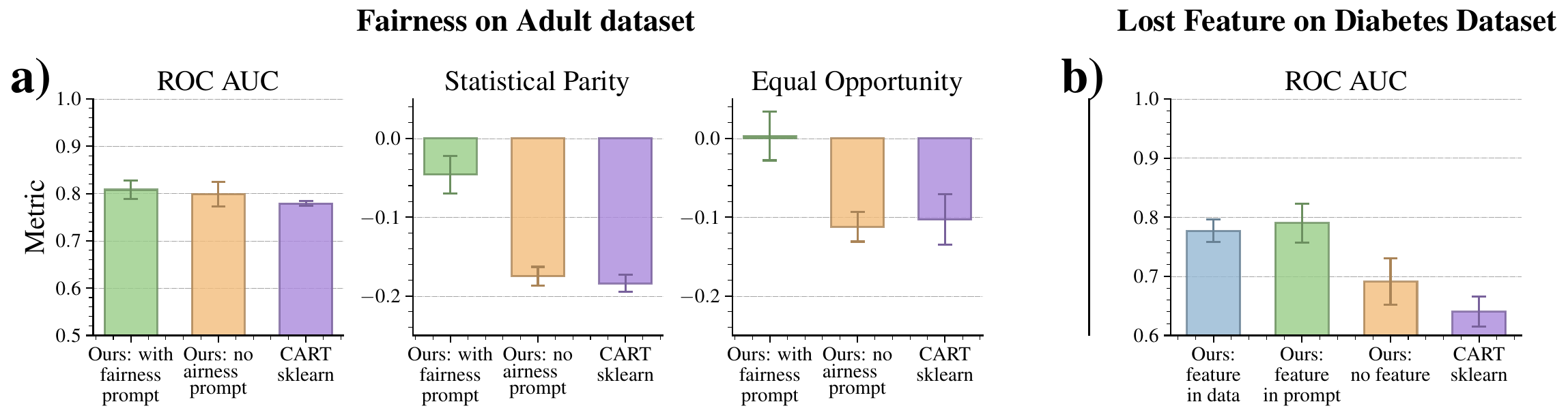}
  \vspace{-15px}
  \caption{\textbf{a)} Fairness evaluation on the Adult dataset across three setups: LLM-built trees with and without the fairness prompt and the \texttt{sklearn} baseline. \textbf{b)} Training experiment on the Diabetes dataset: performance of trees trained with and without access to the \texttt{Glucose} feature. Both experiments use the GPT-5 backbone and the setup from Section~\ref{sect:exp_main_results}.}
  \label{fig:human_input_res}
  \vspace{-5px}
\end{figure*}

As shown in Figure~\ref{fig:human_input_res}~(b), both the \textbf{Feature in prompt} and \textbf{Feature in data} trees achieved similar performance, outperforming trees trained without the feature. This suggests that small datasets, combined with the LLM's prior knowledge, enable effective compensation for missing features.

\vspace{-8pt}
\subsection{Fairness}\label{subsec:fairness}
\vspace{-5pt}
Fairness has become an important concern in machine learning, as predictive models are increasingly deployed in socially sensitive domains~\cite{dwork2012fairness,chouldechova2018_frontiers_fairness,barocas2023fairness}. The research community has proposed several criteria of fairness, such as group-based and individual statistical parity~\cite{dwork2012fairness,mitchell2018prediction}, and practical algorithms for incorporating fairness constraints into classifiers~\cite{agarwal2018reductions,friedler2018_comparative_fairness_interventions}. In our case, we incorporate fairness constraints via the system prompt and tools.

We evaluate on the \texttt{Adult}~\cite{kohavi1996adult} binary classification dataset~\cite{kohavi1996adult}, which is known to contain gender bias\footnote{\texttt{Adult} also has known limitations~\cite{ding2021retiring} that can be important for dedicated fairness studies, but it is sufficient for our simple analysis. See Appendix~\ref{app:fairness:adult} for more details.}. We evaluate fairness in terms of two widely used fairness metrics: \emph{Statistical Parity}~\cite{dwork2012fairness} and \emph{Equal Opportunity}~\cite{equal_opportunity}. If $Y \in \{0,1\}$ is the classification target, $\hat{Y}$ is the model prediction, and $A \in \{a, b\}$ is the protected attribute (gender), then the Statistical Parity Difference (SPD) is:
\[
P(\hat{Y}=1 \mid A=a) - P(\hat{Y}=1 \mid A=b).
\]
The Equal Opportunity Difference (EOD):
\[
P(\hat{Y}=1 \mid A=a, Y=1) - P(\hat{Y}=1 \mid A=b, Y=1).
\]
A perfectly fair model would yield $\text{SPD}=\text{EOD}=0$. Informally, smaller \textit{absolute values} of SPD and EOD correspond to ``fairer'' predictions according to these metrics. We compare the following three method configurations:
\begin{itemize}[leftmargin=*]
    \vspace{-4px}\item \textbf{Ours: with fairness prompt} -- LLM-built tree with an additional prompt (see Appendix~\ref{app:prompts_human_input});
    \vspace{-4px}\item \textbf{Ours: no fairness prompt} -- LLM-built tree without additional fairness-related instructions;
    \vspace{-4px}\item \textbf{CART sklearn} -- decision tree trained using \texttt{Scikit-Learn} with the same tuning protocol.
\end{itemize}

We report our results in Figure~\ref{fig:human_input_res}~(a). To summarize, an informal request to adjust for gender bias results in significantly reduced SPD and EOD. As expected, the bias-adjusted tree also has somewhat lower AUC, but still outperforms a \texttt{sklearn}-based tree. Note that this is a coarse initial exploration of fairness meant to test the versatility of our algorithm. We provide additional experiments including other datasets and experimental details in Appendix~\ref{app:fairness}.

\vspace{-8pt}
\subsection{Monotonic constraints}\label{sec:mono}
\vspace{-5pt}

Many deployed tabular models are expected to respect monotonicity constraints because of domain insights, legal or other reasons. Using simple prompts we can incorporate that requirement into our Talking Trees. To test that we specify a set of features with their expected monotonic directions and evaluate monotonicity using the Normalized Monotonicity Index (NMI) based on ICE~\cite{Goldstein02012015} analysis. For a monotonic feature $i$ with direction $s_i = +1$ (nondecreasing) or $s_i = -1$ (nonincreasing), the NMI is defined as:
\(
\mathrm{NMI} = \frac{B - A}{B + A},
\)
where $A$ is the number of cases with at least one violation and $B$ is the number of cases without violations. The metric ranges from $-1$ to $1$, where $1$ indicates perfect monotonicity. We compare four methods: (1) CART (unconstrained baseline); (2) GPT-5 with monotonicity requested; (3) LightGBM with $n=100$ estimators and monotonic constraints; (4) LightGBM with $n=1$ estimator and monotonic constraints. 

\begin{table}[t]
\centering
\small
\setlength{\tabcolsep}{3pt}
\renewcommand{\arraystretch}{1.05}
\caption{Comparison of Methods by ROC-AUC and NMI. Best NMI per dataset is bolded.}
\label{tab:nmi_results}
\begin{tabular}{@{}lcccccccc@{}}
\toprule
& \multicolumn{2}{c}{\textbf{Heart Disease}} 
& \multicolumn{2}{c}{\textbf{COMPAS}} 
& \multicolumn{2}{c}{\textbf{BlogFeedback}} 
& \multicolumn{2}{c}{\textbf{LoanDefaulter}} \\
\cmidrule(lr){2-3} \cmidrule(lr){4-5} \cmidrule(lr){6-7} \cmidrule(l){8-9}
& ROC-AUC & NMI & ROC-AUC & NMI & RMSE & NMI & ROC-AUC & NMI \\
\midrule
CART &
\shortstack{0.67\\{\tiny\color{gray}$\pm 0.06$}} &
\shortstack{0.83\\{\tiny\color{gray}$\pm 0.14$}} &
\shortstack{0.62\\{\tiny\color{gray}$\pm 0.02$}} &
\shortstack{0.45\\{\tiny\color{gray}$\pm 0.03$}} &
\shortstack{0.21\\{\tiny\color{gray}$\pm 0.00$}} &
\shortstack{0.74\\{\tiny\color{gray}$\pm 0.04$}} &
\shortstack{0.57\\{\tiny\color{gray}$\pm 0.00$}} &
\shortstack{0.00\\{\tiny\color{gray}$\pm 0.05$}} \\
\midrule
GPT-5 &
\shortstack{0.77\\{\tiny\color{gray}$\pm 0.09$}} &
\shortstack{\textbf{1.00}\\{\tiny\color{gray}$\pm 0.00$}} &
\shortstack{0.71\\{\tiny\color{gray}$\pm 0.02$}} &
\shortstack{\textbf{1.00}\\{\tiny\color{gray}$\pm 0.00$}} &
\shortstack{0.15\\{\tiny\color{gray}$\pm 0.00$}} &
\shortstack{0.99\\{\tiny\color{gray}$\pm 0.01$}} &
\shortstack{0.69\\{\tiny\color{gray}$\pm 0.01$}} &
\shortstack{\textbf{1.00}\\{\tiny\color{gray}$\pm 0.00$}} \\
\midrule
LGBM $n=100$ &
\shortstack{\textbf{0.86}\\{\tiny\color{gray}$\pm 0.02$}} &
\shortstack{\textbf{1.00}\\{\tiny\color{gray}$\pm 0.00$}} &
\shortstack{0.72\\{\tiny\color{gray}$\pm 0.01$}} &
\shortstack{\textbf{1.00}\\{\tiny\color{gray}$\pm 0.00$}} &
\shortstack{\textbf{0.15}\\{\tiny\color{gray}$\pm 0.00$}} &
\shortstack{\textbf{1.00}\\{\tiny\color{gray}$\pm 0.00$}} &
\shortstack{\textbf{0.71}\\{\tiny\color{gray}$\pm 0.00$}} &
\shortstack{\textbf{1.00}\\{\tiny\color{gray}$\pm 0.00$}} \\
\midrule
LGBM $n=1$ &
\shortstack{0.79\\{\tiny\color{gray}$\pm 0.03$}} &
\shortstack{\textbf{1.00}\\{\tiny\color{gray}$\pm 0.00$}} &
\shortstack{\textbf{0.72}\\{\tiny\color{gray}$\pm 0.00$}} &
\shortstack{\textbf{1.00}\\{\tiny\color{gray}$\pm 0.00$}} &
\shortstack{0.18\\{\tiny\color{gray}$\pm 0.00$}} &
\shortstack{\textbf{1.00}\\{\tiny\color{gray}$\pm 0.00$}} &
\shortstack{0.70\\{\tiny\color{gray}$\pm 0.00$}} &
\shortstack{\textbf{1.00}\\{\tiny\color{gray}$\pm 0.00$}} \\
\bottomrule
\end{tabular}
\end{table}
Results presented in Table~\ref{tab:nmi_results} suggest that such prompting can steer Talking Trees toward near-perfect NMI across all datasets. Thus our method outperforms CART while remaining competitive with LightGBM, which guarantees perfect NMI. Additional details are presented in Appendix~\ref{app:monotonicity}.

\vspace{-10pt}
\section{Conclusion}\label{sect:conclusion}
\vspace{-8pt}
We have shown that LLM agents can construct and improve decision trees without specialized training, effectively combining tabular data, feature descriptions, and informal domain knowledge. While individual LLM-built trees do not outperform black-box foundation models, they offer a lightweight, interpretable alternative that can incorporate task-specific demands. When combined with other state-of-the-art models, agentic trees show competitive performance while still providing a way for human input. This broadens the role of human input by allowing domain experts to influence model behavior through natural language --- for example, emphasizing fairness or handling missing features.

\textbf{Future work.} Our exploration provides several directions for future research. First, it is worth exploring modifications to traditional trees that would not make sense for CART but can be generated by LLMs, such as complex expressions as splitting criteria or graphs with richer structure. Another question is how best to ensemble agentic decision trees --- while compatible with traditional bagging and boosting, one could also consider LLM-specific ensembling, such as giving each tree a different informal outline. Finally, it is worth exploring principled ways to combine agentic trees with black-box tabular models.

\newpage
\bibliography{bibliography}
\bibliographystyle{unsrt}

\clearpage
\appendix
\onecolumn

\section{Evaluation Setup Details}\label{app:eval_setup}

\subsection{Dataset Information}\label{app:datasets_info}
The main evaluation uses the following datasets (with OpenML task IDs).

\begin{itemize}
  \item \textbf{Fitness} - \texttt{Fitness\_Club} (ID 363671), 1500 samples
  \item \textbf{Marketing} - \texttt{Marketing\_Campaign} (ID 363684), 2240 samples
  \item \textbf{CreditG} - \texttt{credit-g} (ID 363626), 1000 samples
  \item \textbf{Diabetes} - \texttt{diabetes} (ID 363629), 768 samples
  \item \textbf{Customer} - \texttt{Is-this-a-good-customer} (ID 363682), 1723 samples
  \item \textbf{QSARbio} - \texttt{qsar-biodeg} (ID 363696), 1054 samples
  \item \textbf{Hazelnut} - \texttt{hazelnut-spread-contaminant-detection} (ID 363674), 2400 samples
  \item \textbf{Blood} - \texttt{blood-transfusion-service-center} (ID 363621), 748 samples
  \item \textbf{Anneal} - \texttt{anneal} (ID 363614), 898 samples
  \item \textbf{Maternal} - \texttt{maternal\_health\_risk} (ID 363685), 1014 samples
  \item \textbf{Phishing} - \texttt{website\_phishing} (ID 363707), 1353 samples
  \item \textbf{MIC} - \texttt{MIC} (ID 363711), 1699 samples
  \item \textbf{Airfoil} - \texttt{airfoil\_self\_noise} (ID 363612), 1503 samples
  \item \textbf{UsedFiat500} - \texttt{Another-Dataset-on-used-Fiat-500} (ID 363615), 1538 samples
  \item \textbf{Concrete} - \texttt{concrete\_compressive\_strength} (ID 363625), 1030 samples
  \item \textbf{Insurance} - \texttt{healthcare\_insurance\_expenses} (ID 363675), 1338 samples
  \item \textbf{QSARfish} - \texttt{QSAR\_fish\_toxicity} (ID 363698), 907 samples
\end{itemize}

\subsection{Model Hyperparameters}\label{app:model_hyperparams}

For \textbf{XGBoost}, \textbf{Tab-M}, and \textbf{TabPFNv2} we directly take the results from the official TabArena repository \cite{erickson2025tabarena}. However for comparability, we truncate the results to the first five repeats of the first fold, i.e., \texttt{fold = 0, repeat = 0,..., 4}. For \textbf{CART}, we use the Scikit-Learn implementation.

For \textbf{OCTree} \cite{nam2024optimized}, we use GPT-4o as a base LLM model. In the original work, the authors have used LLama-2 7B model, we find that the GPT-4o significantly outperforms the LLama model and crucially outputs consistent and well-formatted trees. We take the tuned hyperparameters of XGBoost from the TabArena. Namely, we note that these parameters are not re-tuned or updated throughout the feature creation iterative loop or the final evaluation, which is in line with the original work.

For \textbf{DeLTA} \cite{ye2025_llm_meeting_decision_trees}, there are 3 components: the random forest which is used for prompting, CART and TabPFN as final decision layers. As a base LLM model similar to the paper, we use GPT-4o. For the random forest, as suggested in the original work, we use the same hyperparameters of -- tuning this could hurt the performance as the prompt was tuned for this structure. For CART, we use max\_depth=5, max\_leaf\_nodes=20 which is the default used for all datasets and evaluations in the original work. For TabPFN, there are essentially no tunable hyperparameters.

\textbf{TAO.} We tune TAO with the wall-clock random-search tuner from \texttt{run\_tuned\_trees.py}, using a 300s budget per dataset split. The initial CART tree hyperparameters are sampled as follows: \texttt{max\_depth} uniformly from 1 to 100, \texttt{min\_samples\_split} uniformly from 2 to 20, \texttt{min\_samples\_leaf} uniformly from 1 to 20, and criterion from \{\texttt{gini}, \texttt{entropy}, \texttt{log\_loss}\} for classification or \{\texttt{squared\_error}, \texttt{friedman\_mse}\} for regression. The TAO-specific regularization is sampled as \(\texttt{reg\_param}=10^u\), where \(u \sim \mathcal{U}[-4,2]\), with \texttt{node\_model=linear}.

\textbf{DPDT.} We tune DPDT with weighted random search under a 300s wall-clock budget per dataset split and select the best validation score, breaking ties by smaller depth and then fewer leaves. The search samples \texttt{max\_depth} from \{5,2,3,4\} with weights \((0.7,0.1,0.1,0.1)\), \texttt{min\_samples\_split} from \{2,3\} with weights \((0.95,0.05)\), \texttt{min\_impurity\_decrease} from \{0,0.01,0.02,0.05\} with weights \((0.85,0.05,0.05,0.05)\), \texttt{min\_samples\_leaf} as a rounded log-uniform integer in [2, 51], \texttt{min\_weight\_fraction\_leaf} from \{0,0.01\} with weights \((0.95,0.05)\), and \texttt{max\_features} from \{\texttt{sqrt}, \texttt{log2}, \texttt{10000}\} with weights \((0.5,0.25,0.25)\), where \texttt{10000} effectively permits all features. We set \texttt{max\_nb\_trees=11}.

\textbf{SAT.} SAT-based tree improvement is classification-only. We use the two-stage tuner from \texttt{sat\_tree\_wrapper.py}: first, grid-search CART on the validation split over \texttt{max\_depth} in \{3,5,7,10,\texttt{None}\}, \texttt{min\_samples\_split} in \{2,5,10\}, \texttt{min\_samples\_leaf} in \{1,5,10\}, and criterion in \{\texttt{gini}, \texttt{entropy}\}; then apply SAT improvement to the top 3 CART trees. The SAT grid uses \texttt{encoding} in \{\texttt{avellaneda}, \texttt{avellaneda2}\} and \texttt{opt\_size} in \{\texttt{True}, \texttt{False}\}, with the \texttt{cadical} solver and a 300s total budget.

\section{Optimal Tree Baseline Details}\label{app:optimal_tree_results}

\textbf{Results and details.} Tables~\ref{tab:optimal_tree_train_val} and~\ref{tab:optimal_tree_complexity_runtime} report the train, validation, and test metrics, hard-label classification results, and tree complexity statistics for the decision-tree baselines. In addition to the agentic tree and the tuned CART baseline, we include three stronger CART-style variants and three non-greedy tree learners. \textbf{CART+AUG} tunes CART together with generic feature augmentations, including additive and multiplicative interactions, sparse feature selection, and PCA. \textbf{CART+AG} trains a tuned CART model on the representation produced by the agent, isolating how much of the gain can be explained by the agent's data preprocessing and feature engineering rather than by the final tree edits. \textbf{TAO}, \textbf{DPDT}, and \textbf{SAT} represent non-greedy or near-optimal tree induction strategies. We include SAT and DPDT in the appendix because their classification outputs are hard labels and further, SAT does not support regression. 

On the test split, the agentic tree improves over tuned CART on 15 of 17 datasets and over TAO on 14 of 17 datasets under the same probabilistic classification and regression metrics. The strongest classical variants do close part of the gap on individual tasks: CART+AG is best on CreditG, Airfoil, and Concrete, while CART+AUG is best on QSARBio. However, these gains are not consistent across the benchmark: the agentic tree still outperforms CART+AUG on 16 of 17 primary test metrics and CART+AG on 12 of 17. The agentic tree also remains strong compared to DPDT and SAT, outperforming DPDT on 19 and SAT on 13 out of 24 classification metrics. On regression, the agentic tree outperforms DPDT on 3 out of 4 datasets. 

\textbf{Complexity and runtime.}
Table~\ref{tab:optimal_tree_complexity_runtime} shows that the agentic trees occupy a useful middle point in the interpretability-performance tradeoff. They average roughly 21 leaves and depth 5.8 across datasets, compared with about 80 leaves for tuned CART, 90 for CART+AG, and 67 for TAO. DPDT is the most compact baseline on average, with about 16 leaves and depth 4.4, but this compactness comes with weaker primary-metric performance in our experiments. Runtime separates the methods into two regimes. CART is essentially instantaneous after preprocessing, while the agentic method, TAO, and DPDT operate on a few-minute scale: the agentic runs average about 295 seconds, and both TAO and DPDT are close to their 300 second tuning budgets. SAT is classification-only and can be highly variable, ranging from about one minute to substantially longer on harder datasets. Overall, these results show that stronger tree learners can improve over greedy CART, but the agentic procedure provides a distinct tradeoff: it obtains strong test performance with a compact single-tree predictor while retaining the ability to incorporate semantic information and user-provided constraints. 

\begin{table*}[t]
\centering
\small
\setlength{\tabcolsep}{2pt}
\renewcommand{\arraystretch}{1.2}
\caption{Performance of decision tree methods across datasets on \textbf{train, validation, and test sets}. \textcolor{gray}{Gray} values are standard deviations over repeats. AUC / log-loss entries are shown as NA when a method only produced hard-label predictions. \textbf{Bold} indicates best performance.}
\label{tab:optimal_tree_train_val}
\begin{adjustbox}{max width=\textwidth}
\begin{tabular}{lccccccccccccccccccccccccccccccccccccccccc}
\toprule
 & \multicolumn{24}{c}{Classification: Binary} & \multicolumn{12}{c}{Multiclass} & \multicolumn{5}{c}{Regression} \\
\cmidrule(lr){2-25}\cmidrule(lr){26-37}\cmidrule(lr){38-42}
& \multicolumn{3}{c}{\makecell{Fitness}} & \multicolumn{3}{c}{\makecell{Marketing}} & \multicolumn{3}{c}{\makecell{CreditG}} & \multicolumn{3}{c}{\makecell{Diabetes}} & \multicolumn{3}{c}{\makecell{Customer}} & \multicolumn{3}{c}{\makecell{QSARBio}} & \multicolumn{3}{c}{\makecell{Hazelnut}} & \multicolumn{3}{c}{\makecell{Blood}} & \multicolumn{3}{c}{\makecell{Anneal}} & \multicolumn{3}{c}{\makecell{Maternal}} & \multicolumn{3}{c}{\makecell{Phishing}} & \multicolumn{3}{c}{\makecell{MIC}} & \makecell{Airfoil} & \makecell{Fiat500} & \makecell{Concrete} & \makecell{Insurance} & \makecell{QSARFish} \\
& {\tiny AUC} & {\tiny F1} & {\tiny Acc} & {\tiny AUC} & {\tiny F1} & {\tiny Acc} & {\tiny AUC} & {\tiny F1} & {\tiny Acc} & {\tiny AUC} & {\tiny F1} & {\tiny Acc} & {\tiny AUC} & {\tiny F1} & {\tiny Acc} & {\tiny AUC} & {\tiny F1} & {\tiny Acc} & {\tiny AUC} & {\tiny F1} & {\tiny Acc} & {\tiny AUC} & {\tiny F1} & {\tiny Acc} & {\tiny LL} & {\tiny F1} & {\tiny Acc} & {\tiny LL} & {\tiny F1} & {\tiny Acc} & {\tiny LL} & {\tiny F1} & {\tiny Acc} & {\tiny LL} & {\tiny F1} & {\tiny Acc} & {\tiny RMSE} & {\tiny RMSE} & {\tiny RMSE} & {\tiny RMSE} & {\tiny RMSE} \\
\midrule
\multicolumn{42}{l}{\textit{\textbf{Train}}} \\
\bottomrule
\end{tabular}
\end{adjustbox}
\end{table*}

\begin{table*}[t]
\centering
\small
\setlength{\tabcolsep}{2pt}
\renewcommand{\arraystretch}{1.2}
\caption{Tree complexity and runtime across datasets. \textcolor{gray}{Gray} values are standard deviations over repeats. \textbf{Bold} indicates smallest value.}
\label{tab:optimal_tree_complexity_runtime}
\begin{adjustbox}{max width=\textwidth}
\begin{tabular}{lccccccccccccccccccccccccccccccccccccccccccccccccccc}
\toprule
& \multicolumn{3}{c}{\makecell{Fitness}} & \multicolumn{3}{c}{\makecell{Marketing}} & \multicolumn{3}{c}{\makecell{CreditG}} & \multicolumn{3}{c}{\makecell{Diabetes}} & \multicolumn{3}{c}{\makecell{Customer}} & \multicolumn{3}{c}{\makecell{QSARBio}} & \multicolumn{3}{c}{\makecell{Hazelnut}} & \multicolumn{3}{c}{\makecell{Blood}} & \multicolumn{3}{c}{\makecell{Anneal}} & \multicolumn{3}{c}{\makecell{Maternal}} & \multicolumn{3}{c}{\makecell{Phishing}} & \multicolumn{3}{c}{\makecell{MIC}} & \multicolumn{3}{c}{\makecell{Airfoil}} & \multicolumn{3}{c}{\makecell{Fiat500}} & \multicolumn{3}{c}{\makecell{Concrete}} & \multicolumn{3}{c}{\makecell{Insurance}} & \multicolumn{3}{c}{\makecell{QSARFish}} \\
\cmidrule(lr){2-4}\cmidrule(lr){5-7}\cmidrule(lr){8-10}\cmidrule(lr){11-13}\cmidrule(lr){14-16}\cmidrule(lr){17-19}\cmidrule(lr){20-22}\cmidrule(lr){23-25}\cmidrule(lr){26-28}\cmidrule(lr){29-31}\cmidrule(lr){32-34}\cmidrule(lr){35-37}\cmidrule(lr){38-40}\cmidrule(lr){41-43}\cmidrule(lr){44-46}\cmidrule(lr){47-49}\cmidrule(lr){50-52}
& {\tiny Lvs} & {\tiny Dep} & {\tiny Sec} & {\tiny Lvs} & {\tiny Dep} & {\tiny Sec} & {\tiny Lvs} & {\tiny Dep} & {\tiny Sec} & {\tiny Lvs} & {\tiny Dep} & {\tiny Sec} & {\tiny Lvs} & {\tiny Dep} & {\tiny Sec} & {\tiny Lvs} & {\tiny Dep} & {\tiny Sec} & {\tiny Lvs} & {\tiny Dep} & {\tiny Sec} & {\tiny Lvs} & {\tiny Dep} & {\tiny Sec} & {\tiny Lvs} & {\tiny Dep} & {\tiny Sec} & {\tiny Lvs} & {\tiny Dep} & {\tiny Sec} & {\tiny Lvs} & {\tiny Dep} & {\tiny Sec} & {\tiny Lvs} & {\tiny Dep} & {\tiny Sec} & {\tiny Lvs} & {\tiny Dep} & {\tiny Sec} & {\tiny Lvs} & {\tiny Dep} & {\tiny Sec} & {\tiny Lvs} & {\tiny Dep} & {\tiny Sec} & {\tiny Lvs} & {\tiny Dep} & {\tiny Sec} & {\tiny Lvs} & {\tiny Dep} & {\tiny Sec} \\
\bottomrule
\end{tabular}
\end{adjustbox}
\end{table*}

\section{Agent Analysis}
\label{app:agent-analysis}

\begin{figure*}[t]
    \centering
    \includegraphics[width=0.9\linewidth]{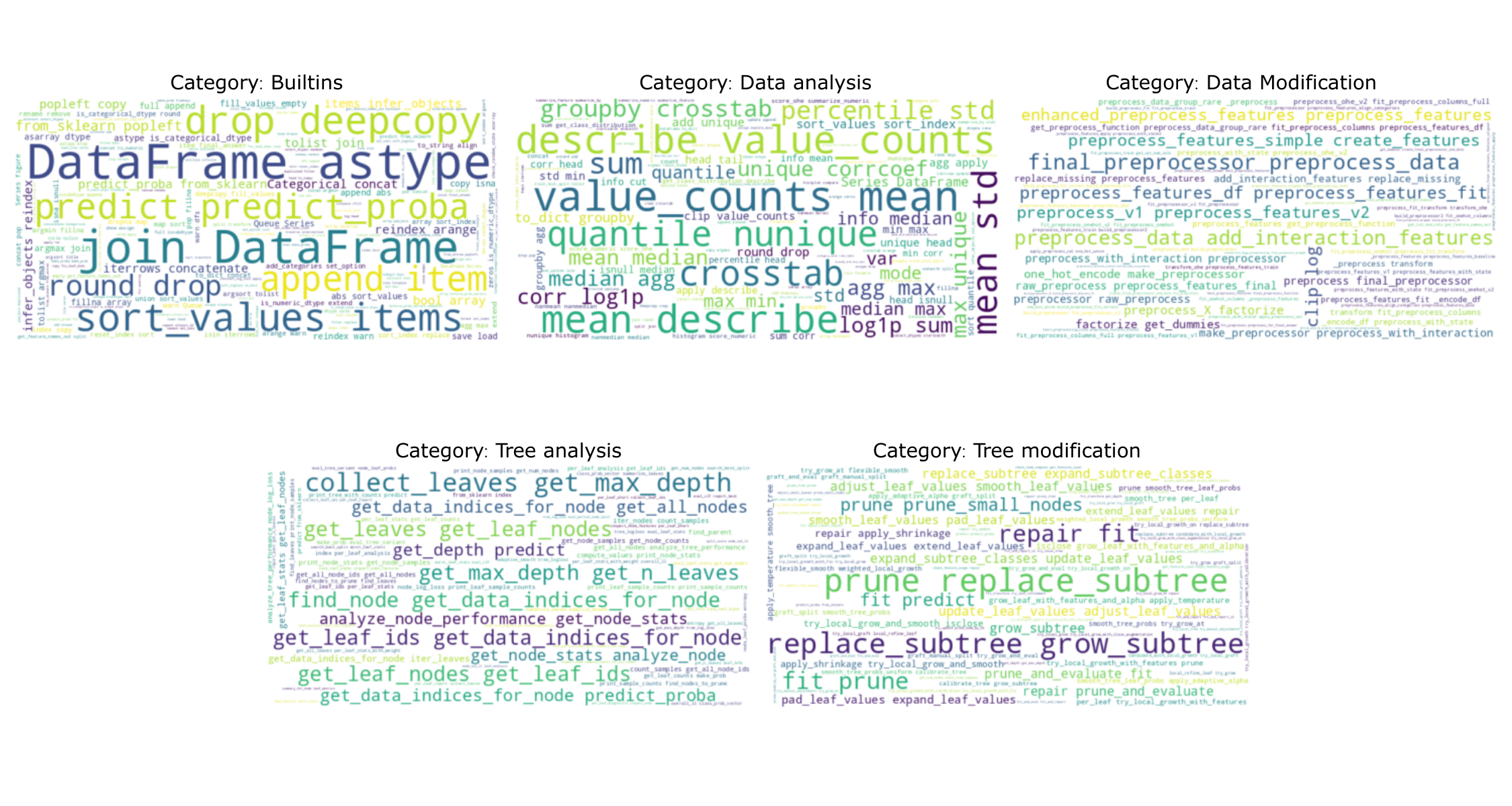}
    \caption{Word cloud of function calls by category.}
    \label{fig:func-wordcloud}
\end{figure*}

As we highlight in Section~\ref{sect:exp_analysis}, we analyze agent's behavior by quantifying its function calling. Specifically, we group its function calls into five distinct categories. These are delineated by their target (the model vs. the data) and their purpose (modification vs. analysis). The categories are: 1) \textbf{tree modification}, which encompasses all actions that train or alter the decision tree's structure, such as pruning or setting hyperparameters; 2) \textbf{tree analysis}, which involves introspecting a trained tree to understand its logic, for example by visualizing decision paths or feature importances; 3) \textbf{feature engineering}, which covers any pre-processing or transformation of input data before it is used by the model; 4) \textbf{general data analysis}, which includes dataset-level exploration to understand its intrinsic properties, such as distributions or correlations, independent of any specific model; 5) \textbf{builtins}, which are infrastructural functions for tasks like I/O or logging that support the workflow but do not manipulate the data or model.

Our analytical procedure is as follows. First, we parse the agent's execution logs using regular expressions to extract all function calls, aggregated separately for each model-dataset combination. Next, we use an LLM (Kimi-2) to classify each unique function into one of the five predefined categories by providing it with the function name and the category descriptions. We provide the classifier prompt below. To visualize the most representative functions within each category, we generate a word cloud for each group, as shown in Figure~\ref{fig:func-wordcloud}. The size of each function in the cloud corresponds to its total call frequency, and a qualitative review confirms the semantic coherence of the automated categorization.

\begin{tcolorbox}[
    colframe=black!70,
    colback=gray!5,
    title=Prompt for Agent Call Classification,
    enhanced,
    breakable,
    skin first=enhanced,
    skin middle=enhanced,
    skin last=enhanced,]
\small\ttfamily
Help categorize the following functions into the following categories:

1. Tree modifications: `tree\_mod` (key)

2. Tree analysis, visualization and debugging: `tree\_eda` (key)

3. General feature engineering and transformations: `feat\_engineering` (key)

4. General exploratory data analysis: `eda` (key)

5. Builtins: `builtins` (key)

Category descriptions:

1. Tree modifications: any operation that changes or trains the tree. Keywords: DecisionTreeClassifier, DecisionTreeRegressor, min\_samples\_split, max\_depth, prune, replace\_subtree, grow\_subtree, repair, min\_samples\_leaf, max\_features, min\_impurity\_decrease, min\_weight\_fraction\_leaf, ccp\_alpha, max\_leaf\_nodes, min\_samples\_leaf, min\_samples\_split, max\_depth, max\_features, max\_leaf\_nodes, min\_impurity\_decrease, min\_samples\_leaf, min\_samples\_split, min\_weight\_fraction\_leaf, random\_state, ccp\_alpha.

2. Tree analysis: introspection of trained tree(s) such as paths, leaves, importances, surrogate views, and plots. Keywords: is\_leaf, get\_data\_indices\_for\_node, print, decision\_path, feature\_importances\_, export\_graphviz.

3. General feature engineering and transformations: any input transformation before training/inference. Independent of a specific trained tree. Keywords: OneHotEncoder, PolynomialFeatures, clip, log1p, sign, datetime64.

4. General exploratory data analysis: dataset-level profiling and exploration not tied to a specific model such as distributions, correlations, missingness, leakage checks, class balance. Keyword: percentile, std, mean, tsne, umap, train\_test\_split.

5. Builtins: infrastructure that doesn’t change data or models and isn’t analysis such as I/O, seeding, logging, timing, config, small data wrangling helpers. Keywords:  check\_random\_state, asarray, dtype, save, load.

As an output provide a json-formattable dictionary of the form: {{category: [function1, function2]}}.
Specifcially, your answer will be directly plugged in to the `json.loads(YOUR\_ANSWER)`. Please, respect the formatting.

Here are the functions: `{set(defined\_functions)}`.
\end{tcolorbox}

\section{Human Input Prompts}
\label{app:prompts_human_input}

The full prompts for all setups are available in our repository.
Below we show the exact additional instructions we provided to the LLM in our human input experiments. Each prompt is inserted as an additional text to the base setup and is designed to inject domain knowledge or constraints that a standard execution would not capture.

Below directions were  used in the fairness experiment (Section~\ref{subsec:fairness}) to explicitly guide the LLM toward balancing outcomes across gender groups in the \texttt{Adult} and \texttt{SchoolPerformance} datasets.
\begin{tcolorbox}[
    colframe=black!70,
    colback=gray!5,
    title=Prompt for Fairness Human Input,
    enhanced,
    breakable,
    skin first=enhanced,
    skin middle=enhanced,
    skin last=enhanced,]
\small\ttfamily
You should prioritize creating model that as gender-neutral as possible even if it harms metrics.
\end{tcolorbox}

The next instruction set was used in the lost feature experiment (Section~\ref{subsec:lost_feature}) to test whether the LLM could integrate prior knowledge about the missing variable into the tree design.

\begin{tcolorbox}[
    colframe=black!70,
    colback=gray!5,
    title=Prompt for Lost Feature Human Input,
    enhanced,
    breakable,
    skin first=enhanced,
    skin middle=enhanced,
    skin last=enhanced,]
\small\ttfamily
We forgot to collect `Glucose` data in the training set. This feature is actually very important for predicting diabetes, but you cannot see it in train. Your task is to create a decision tree that may rely on this feature in the test set, even though you have no Glucose data available during training.
\end{tcolorbox}

\section{Tree Construction Examples}
In this section we present example reasoning traces produced during the tree construction process. These traces illustrate the step-by-step decisions made by the LLM while building or modifying decision trees under different constraints. For all experiments we used GPT-5 as the underlying model. In several examples code has been removed for readability. The following excerpts are shown to highlight representative reasoning patterns rather than complete outputs.

\subsection{Main Experiment Examples}
As an illustration of the LLM's workflow, we provide a reasoning trace consisting solely of the model’s intermediate thoughts during a single run of tree construction. No code is shown, but the trace remains internally consistent and reflects the step-by-step process followed in that run.

\begin{tcolorbox}[
    colframe=black!70,
    colback=gray!5,
    title=Trace example on \texttt{Fitness} dataset (redacted),
    enhanced,
    breakable,
    skin first=enhanced,
    skin middle=enhanced,
    skin last=enhanced,]
\small\ttfamily
- Thought: I will start by inspecting the raw data: shapes, NaN counts, class balance, and unique values for categorical features. Hypothesis: there are NaNs in weight, and potentially unseen categories in validation (e.g., category "-"), so I will need imputation and robust one-hot encoding.

- Thought: I will design a robust preprocessing pipeline: impute missing weight with the train median and add a missing indicator; one-hot encode categorical features; ensure consistent columns between train/val. Then I will train a few sensible baseline DecisionTreeClassifier variants (small set of hyperparameters, with class\_weight balanced) and pick the best by validation AUC. I will convert it to editable\_tree.Tree and print structure to form hypotheses for manual edits.

- Thought: I will inspect the current best tree’s leaves and internal nodes by computing how many train/val samples route to each node and the mean target (attendance rate) within those nodes. Hypothesis: some large leaves may still be heterogeneous (e.g., split further by days\_before or time), offering room for local grafting/growth to improve AUC. I will list all nodes, identify leaves, print their stats, then propose targeted edits on the biggest/most heterogeneous leaves.

- Thought: Hypothesis: within large leaves, days\_before, time, or category may provide additional separation to improve AUC. I will select the top-2 largest leaves and try growing a small local subtree (max\_depth=2, min\_samples\_leaf=20) using a focused set of features, then graft it, evaluate AUC, and keep changes only if validation AUC improves.

- Thought: I will analyze univariate relationships to guide targeted edits. Hypothesis: days\_before and time or category may add predictive power, especially within the lower-membership region (months\_as\_member <= 15.5). I’ll compute univariate AUCs and attendance means by bins/levels. Then I’ll try local growth at key internal nodes (ids 1, 10, 11, 13), training small subtrees on the routed subsets and grafting them. I’ll keep any graft that improves validation AUC.

- Thought: I’ll add a few interpretable engineered features that reflect plausible behavior: weekend class indicator (Sat/Sun), early booking (>=10 and >=7 days), high weight (>=85 kg), new member (<=12 months), and intense class (HIIT or Cycling). Hypothesis: these features will help the tree form cleaner splits with the same complexity and may improve AUC. Then I’ll retrain a tree with the previously best hyperparameters, inspect AUC, and if better, convert to editable\_tree for potential small local refinements (focused threshold tuning limited to weight-only splits on the right branch).

\end{tcolorbox}

Next, we present several selected examples of thought–code pairs extracted out of their full experimental context. These examples serve to illustrate what the agent is actually doing to achieve the stated in thoughts goals in constructing decision trees.

\begin{tcolorbox}[
    colframe=black!70,
    colback=gray!5,
    title=Thought-Code Pair example of local graft,
    enhanced,
    breakable,
    skin first=enhanced,
    skin middle=enhanced,
    skin last=enhanced,]
\small\ttfamily
Thought:
I will select the top-2 largest leaves and try growing a small local subtree (max\_depth=2, min\_samples\_leaf=20) using a focused set of features, then graft it.

Code:
\begin{minted}[breaklines]{python}
def try_local_grow(tree, node_id, Xtr, ytr, Xva, yva, feat_cols, max_depth=2, min_samples_leaf=20):
    ids_tr = tree.get_data_indices_for_node(node_id, Xtr)
    ids_va = tree.get_data_indices_for_node(node_id, Xva)
    if len(ids_tr) < 2*min_samples_leaf:
        return None, None, None  # not enough data to grow
    # Fit a small subtree using a subset of features
    clf = DecisionTreeClassifier(random_state=0, class_weight="balanced",
                                 max_depth=max_depth, min_samples_leaf=min_samples_leaf)
    clf.fit(Xtr.iloc[ids_tr][feat_cols], ytr[ids_tr])
    # Evaluate partial AUC for diagnostics
    part_auc = roc_auc_score(yva[ids_va], clf.predict_proba(Xva.iloc[ids_va][feat_cols])[:,1]) if
    len(np.unique(yva[ids_va]))>1 else np.nan
    # Graft into a copy of the tree
    new_tree = deepcopy(tree)
    new_tree.replace_subtree(node_id, Tree.from_sklearn(clf))
    # Evaluate global AUC
    val_auc = roc_auc_score(yva, new_tree.predict(Xva))
    train_auc = roc_auc_score(ytr, new_tree.predict(Xtr))
    return new_tree, train_auc, val_auc

# Try local growth on top-2 leaves
candidates = []
for lid, ntr, nva, mtr, mva in leaf_stats_sorted[:2]:
    print(f"\nAttempt local growth at leaf id={lid} (train n={ntr}, val n={nva})")
    new_tree, tr_auc, va_auc = try_local_grow(
        etree, lid, Xtr, y_train, Xva, y_val,
        focused_feats, max_depth=2, min_samples_leaf=20
    )
    if new_tree is not None:
        print(f" Local growth result -> Train AUC={tr_auc:.4f}, Val AUC={va_auc:.4f}")
        candidates.append((va_auc, tr_auc, lid, new_tree))
    else:
        print(" Not enough samples to grow here.")
\end{minted}
\end{tcolorbox}

\begin{tcolorbox}[
    colframe=black!70,
    colback=gray!5,
    title=Thought-Code Pair example of Leaf label smoothing,
    enhanced,
    breakable,
    skin first=enhanced,
    skin middle=enhanced,
    skin last=enhanced,]
\small\ttfamily
Thought: I’ll try three conservative, targeted tweaks and keep any that improve or match validation AUC:

- Laplace-smooth leaf probabilities to reduce overconfident small leaves. Hypothesis: small leaves on the right branch could be slightly overconfident; smoothing might improve ranking.

Code:
\begin{minted}[breaklines]{python}
def laplace_smooth_tree(tree_in, Xproc, y, alpha=1.0):
    t = deepcopy(tree_in)
    for leaf in get_leaf_nodes(t):
        idx = t.get_data_indices_for_node(leaf.id, Xproc)
        n = len(idx)
        if n == 0:
            continue
        pos = float(y[idx].sum())
        leaf.value = (pos + alpha) / (n + 2*alpha)
    return t

sm_tree = laplace_smooth_tree(best_tree, Xtr, y_train, alpha=1.0)
sm_tr_auc = roc_auc_score(y_train, sm_tree.predict(Xtr))
sm_va_auc = roc_auc_score(y_val, sm_tree.predict(Xva))
print(f"Laplace(alpha=1): train AUC={sm_tr_auc:.4f}, val AUC={sm_va_auc:.4f}")
candidate_tree = deepcopy(best_tree)
cand_tr_auc, cand_va_auc = base_tr_auc, base_va_auc

if sm_va_auc >= cand_va_auc:
    candidate_tree = deepcopy(sm_tree)
    cand_tr_auc, cand_va_auc = sm_tr_auc, sm_va_auc
    print("Adopted Laplace smoothing as candidate.")
\end{minted}
\end{tcolorbox}

\subsection{TabPFN Correction Examples}
Here we present examples of thought–code pairs extracted from the traces of the TabPFN Correction experiments.
\begin{tcolorbox}[
    colframe=black!70,
    colback=gray!5,
    title=TabPFN Correction Thought-Code pair example,
    enhanced,
    breakable,
    skin first=enhanced,
    skin middle=enhanced,
    skin last=enhanced,]
\small\ttfamily
Thought: The numeric stump is close but slightly below baseline. Hypothesis: adding a second split on credit\_history on the high amount\_per\_month side can capture stable residual bias (critical account slightly underpredicted, existing paid duly slightly overpredicted). I will:

- Manually build a 2-level tree: root split on amount\_per\_month; right child splits on credit\_history (critical vs existing duly vs others).

Code:
\begin{minted}[breaklines]{python}
# Helper to build a manual tree given threshold and leaf values
def build_tree(thresh, v_low, v_crit, v_exist, v_other):
    # Right branch: first check critical account, else check existing duly, else other

    # One-hot column names (ensure they exist; if not, they will be zeros)
    crit_col = "credit_history_critical account / other credits existing"
    exist_col = "credit_history_existing credits paid duly till now"

    # Construct right subtree
    right = Tree(
            id=10, feature=crit_col, threshold=0.5,
            left=Tree(  # not critical
            id=11, feature=exist_col, threshold=0.5,
            left=Tree(id=12, value=float(v_other)),   # not existing duly -> "other" categories
            right=Tree(id=13, value=float(v_exist))   # existing duly
        ),
        right=Tree(id=14, value=float(v_crit))         # critical account
    )
    root = Tree(id=0, feature="amount_per_month", threshold=float(thresh),
        left=Tree(id=1, value=float(v_low)),
        right=right)
    return root

for t in thresholds:
    m_left, m_crit, m_exist, m_other = residual_means(t)
    for eta in etas:
        tree_cand = build_tree(t, eta*m_left, eta*m_crit, eta*m_exist, eta*m_other)

        # Evaluate AUCs when added to baseline
        ytr_pred = y_train_baseline + tree_cand.predict(Xtr_ch)
        yva_pred = y_val_baseline + tree_cand.predict(Xva_ch)
        tr_auc = roc_auc_score(y_train, ytr_pred)
        va_auc = roc_auc_score(y_val, yva_pred)
        results.append((t, eta, m_left, m_crit, m_exist, m_other, tr_auc, va_auc, tree_cand))
        print(f"MANUAL depth2 t={t}, eta={eta}: Train AUC={tr_auc:.6f} Val AUC={va_auc:.6f}")
\end{minted}

\end{tcolorbox}

\subsection{Fairness Example}

To better illustrate how the LLM incorporates fairness instructions into the tree-building process, we show below a redacted internal reasoning trace on the \texttt{Adult} and \texttt{SchoolPerformance} datasets.

\begin{tcolorbox}[
    colframe=black!70,
    colback=gray!5,
    title=Fairness trace example on \texttt{Adult} dataset (redacted),
    enhanced,
    breakable,
    skin first=enhanced,
    skin middle=enhanced,
    skin last=enhanced,]
\small\ttfamily
Thought: I will inspect the raw data to understand dtypes, missing values , class balance, and categorical levels.

Thought: The current tree relies heavily on marital-status at the root, which is strongly correlated with gender outcomes via roles like Husband/Wife (even though “relationship” itself is excluded). To improve gender neutrality, I will first drop the relationship feature, then explicitly retrain while excluding both relationship and marital-status to reduce gender proxies. I will iteratively adjust depth/regularization to avoid rediscovering proxy splits.

Thought: I will try to reduce gender disparity without using sex as a feature by reweighting the training samples to give higher weight to female records during training. Hypothesis: upweighting females will increase their predicted scores and narrow the positive-rate gap and AUC gap between sexes. I may also try dropping occupation (a potential proxy) if needed.

Thought: I will explore additional proxy feature removal to improve gender neutrality without using the sex feature. Hypothesis: dropping occupation (and possibly workclass and race) will reduce gender proxy signals in the splits, narrowing the male/female disparities in AUC and positive rate, at some cost to overall AUC. I will try several drop sets (keeping numeric features, education, native-country).

Thought: Aim: reduce gender proxy effects further by removing numeric features that may indirectly encode gender patterns, especially hours-per-week (known to differ by sex). Hypothesis: dropping hours-per-week (and possibly fnlwgt) will narrow positive-rate gaps and AUC gaps by sex, at some cost to overall AUC.

Thought: I will target fairness by increasing predicted probabilities at leaves that predominantly contain female samples, without using sex as an input feature. Hypothesis: modestly increasing leaf values for female-heavy leaves will raise female positive rates more than male (since more females route to those leaves), narrowing the gap while keeping ROC AUC nearly unchanged. Plan:

- Identify leaves and compute, for each leaf, the val-sample counts and female share.

- For candidate uplift deltas (0.02, 0.05, 0.08), adjust leaf values where female share >= 0.6 (and also try >= 0.55) with clipping to [0,1].

- Evaluate val AUC and fairness metrics; choose the best tree by fairness score, tie-breaking by val AUC.

- Keep everything reproducible and print diagnostics.
\end{tcolorbox}

\begin{tcolorbox}[
    colframe=black!70,
    colback=gray!5,
    title=Fairness trace example on \texttt{SchoolPerformance} dataset (redacted),
    enhanced,
    breakable,
    skin first=enhanced,
    skin middle=enhanced,
    skin last=enhanced,]
\small\ttfamily
Thought: I want to identify which features likely correspond to 3-category one-hot encodings (e.g., Gender and Family\_Income, both with 3 levels) so I can exclude them for fairness. Hypothesis: any triple of binary features that sums to exactly 1 for every row (in both train and val) is a 3-category one-hot group. I will:

- Scan all column triples among binary features (n\_unique=2) and test the exact-one condition across both splits.

- Print detected triples and their per-class mean y to see their relation to outcomes.

- Drop all columns in detected triples to reduce gender dependence (this will also remove Family\_Income, which is demographic; acceptable per fairness priority).

- Retrain a comparable tree and compare AUC.

- Inspect the new tree to ensure it avoids those features.

\end{tcolorbox}

\section{Fairness Experimental Details}
\label{app:fairness}

\subsection{Adult Experiments}\label{app:fairness:adult}

We chose the \texttt{Adult} dataset despite its known limitations \cite{ding2021retiring} such as class imbalance, demographic biases, and potential label noise, because it is a widely used benchmark in fairness research. Its prevalence enables direct comparison of our fairness metrics with prior work, and its protected attribute structure (e.g., gender) is well suited for controlled experiments on fairness.

The Fairness Section~\ref{subsec:fairness} in the main text reports a CART baseline (``CART tuned'') that tunes \texttt{min\_samples\_split} via Optuna's Treestructured Parzen Estimator (TPE) sampler with 5 trials, leaving every other CART hyperparameter at its  \texttt{scikit-Learn} default. While this matches the typical practitioner setup, it does not place CART on the accuracy--fairness frontier. In this section we tune CART independently for each of the three target metrics (ROC AUC, statistical parity, equal opportunity), using the same Optuna TPE sampler but with a budget of 1000 trials over an extended hyperparameter space.

For each of the five folds used, we run three independent Optuna studies, one per target metric, each with 1000 trials. Total budget: $5 \times 3 \times 1000 = 15{,}000$ tree fits. Each trial samples a configuration of \texttt{sklearn.tree.DecisionTreeClassifier} from the extended space in Table~\ref{tab:cart_search_space}. We chose it to grant the classical baseline large reasonable advantage.

\begin{table}[b]
  \centering
  \caption{Extended search space used by Optuna~TPE for additional fairness experiments in Section~\ref{app:fairness}.}
  \label{tab:cart_search_space}
  \begin{tabular}{ll}
    \toprule
    Hyperparameter & Range \\
    \midrule
    \texttt{max\_depth}             & $\{4, 6, 8, 10, 12, 15, 20, 25, 30, \mathrm{None}\}$ \\
    \texttt{min\_samples\_split}    & integer in $[2, 500]$ \\
    \texttt{min\_samples\_leaf}     & integer in $[1, 200]$ \\
    \texttt{max\_leaf\_nodes}       & $\{\mathrm{None}, 50, 100, 200, 500, 1000, 2000\}$ \\
    \texttt{min\_impurity\_decrease}& log-uniform in $[10^{-7}, 10^{-2}]$ \\
    \texttt{criterion}              & $\{\mathrm{gini}, \mathrm{entropy}, \mathrm{log\_loss}\}$ \\
    \texttt{splitter}               & $\{\mathrm{best}, \mathrm{random}\}$ \\
    \texttt{max\_features}          & $\{\mathrm{None}, \mathrm{sqrt}, \mathrm{log2}, 0.5, 0.7, 0.9\}$ \\
    \texttt{ccp\_alpha}             & log-uniform in $[10^{-6}, 10^{-2}]$ \\
    \texttt{class\_weight}          & $\{\mathrm{None}, \mathrm{balanced}\}$ \\
    \bottomrule
  \end{tabular}
\end{table}

The objectives are
\begin{align}
  \mathcal{L}_{\mathrm{AUC}}(\theta) &= -\,\mathrm{AUC}(\theta), \\
  \mathcal{L}_{\mathrm{SP}}(\theta)  &= |\mathrm{SP}(\theta)| + \lambda \cdot \mathbf{1}[\mathrm{AUC}(\theta) < \tau], \\
  \mathcal{L}_{\mathrm{EO}}(\theta)  &= |\mathrm{EO}(\theta)| + \lambda \cdot \mathbf{1}[\mathrm{AUC}(\theta) < \tau],
\end{align}
with AUC floor $\tau = 0.70$ and penalty $\lambda = 10$. Best trees are selected by the corresponding study's optimum.
If we drop the AUC penalty and instruct TPE to minimize $|\mathrm{SP}|$ or $|\mathrm{EO}|$ alone, every split converges within a few hundred trials to a tree that predicts the negative class for every input. Such a tree has $P(\hat y = 1 \mid \mathrm{group}) = 0$ identically across groups, so $\mathrm{SP} = \mathrm{EO} = 0$ by construction. AUC degrades to $0.500$. A working fairness objective therefore requires either an explicit accuracy constraint (as in our $\tau = 0.70$ penalty), a Lagrangian combination $\mathcal{L} = |\mathrm{SP}| - \alpha \cdot \mathrm{AUC}$, or a multi-objective formulation that returns a Pareto front rather than a single optimum. Each of these choices is itself non-trivial.

\begin{figure*}[t]
\includegraphics[width=\textwidth]{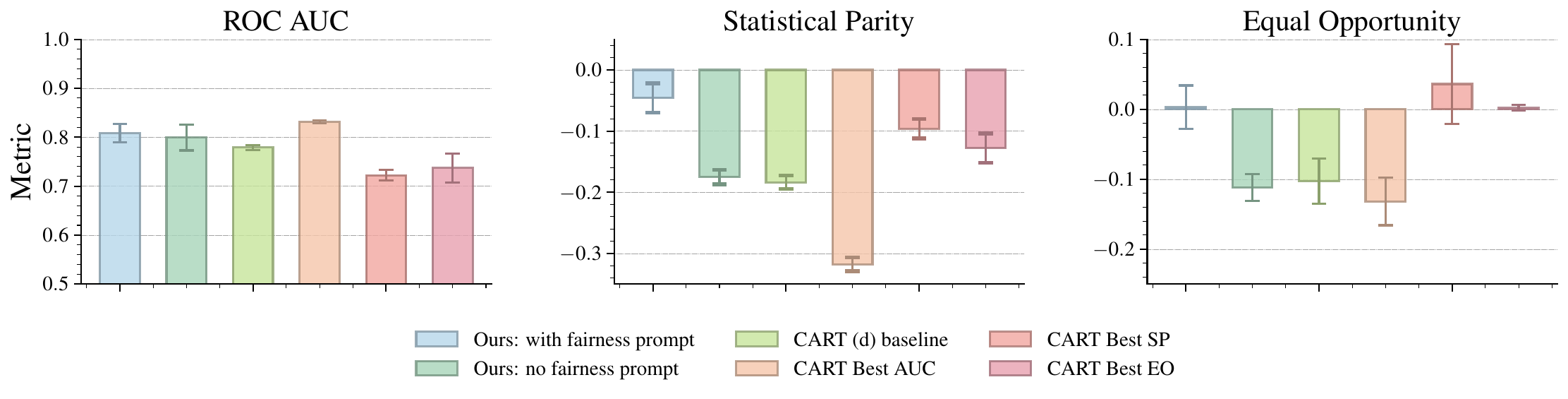}
\caption{Comparison across six experimental setups, including three additional CART configurations obtained via 1000-sample hyperparameter search via OptunaTPE.}
\label{fig:fairness_bo1000}
\end{figure*}

Figure~\ref{fig:fairness_bo1000} compares six configurations across ROC AUC, statistical parity, and equal opportunity: TalkingTrees with and without the fairness prompt, the original \texttt{min\_samples\_split}-tuned CART baseline, and the three Optuna-tuned CART variants (``Best AUC'', ``Best SP'', ``Best EO''). Three observations are immediate. First, CART optimized purely for AUC wins on accuracy by a clear margin (0.831 vs.\ TalkingTrees' 0.808) but at the cost of the largest observed disparity on both fairness metrics ($|\mathrm{SP}| = 0.318$, $|\mathrm{EO}| = 0.132$). Second, CART tuned for fairness reaches $|\mathrm{SP}| = 0.096$ and $|\mathrm{EO}| = 0.002$ only by sacrificing $0.09$--$0.11$ of AUC, landing well below TalkingTrees with the fairness prompt. Third, no Optuna setup simultaneously matches TalkingTrees on both axes: the prompted
model dominates Best-SP and Best-EO on AUC, and dominates Best-AUC on every fairness metric.

In the end, the classical pipeline can match TalkingTrees on a single axis, but matching the joint behavior requires a researcher to (i) design a fairness-respecting objective, (ii) calibrate at least one accuracy threshold, and (iii) invest a non-trivial compute budget for hyperparameter search. TalkingTrees achieves a comparable point on the frontier with no objective design and minimal compute, by translating the user's intent into the construction process directly.

\subsection{SchoolPerformance Experiments}\label{app:fairness:SchoolPerformance}

We evaluate our Talking Trees using the ABCFair benchmark~\cite{defrance2024abcfair} on the \texttt{SchoolPerformance} prediction task~\cite{lenders2023real}. This dataset is particularly interesting as it contains dual labels: biased labels from human labelers based on student features, and unbiased labels reflecting actual outcomes. Under the ABCFair protocol, models are trained on the biased labels and evaluated on the unbiased ones. The authors of ABCFair observe a synergy effect, where fairer models tend to achieve improved performance.

ABCFair summarizes fairness using a \emph{violation score}, which measures how much the group statistic $S_g$ deviates from the population mean $S$ and takes the maximum over the groups:
\[
\mathrm{Viol}(S) := \max_{g \in \mathcal{G}} \left| \frac{S_g}{S} - 1 \right|.
\]
Intuitively, this corresponds to the worst-off group performance. A violation of zero means all groups receive positive predictions at the same rate; higher values indicate greater disparity. Demographic parity and equal opportunity can be recovered from this metric by choosing the appropriate statistic~\cite{defrance2024abcfair}.

Our results in Table~\ref{tab:abcfair_full_exp_values} show that injecting fairness at the prompt level improves predictive accuracy for our tree model compared to the default prompt. We also observe that 5 out of 7 fairness violations decrease, especially demographic parity and equal opportunity. Overall, this experiment indicates that prompt-level injection can help mitigate biases. At the same time, the complete metric breakdown shows an important caveat: spillover bias to other unprotected attributes can occur if not all relevant protected features are specified.

\begin{table*}[t]
\centering
\scriptsize
\setlength{\tabcolsep}{5pt}
\renewcommand{\arraystretch}{1.1}
\begin{tabular}{l l c c }
\toprule
Category & Metric & {Default prompt} & {Fairness prompt} \\
\midrule

\multicolumn{4}{l}{\textbf{Overall (not audited)}} \\
Perf. & Accuracy $\uparrow$                 & 0.8314  & 0.8547 \\
Perf. & AUROC $\uparrow$                    & 0.7110  & 0.7089 \\
Perf. & F1 $\uparrow$                       & 0.8990  & 0.9147 \\
Time  & Training time (s) $\downarrow$      & 243.283 & 360.833 \\
Time  & Total runtime (s) $\downarrow$      & 247     & 363 \\
\midrule

\multicolumn{4}{l}{\textbf{Binary audit (violation $\downarrow$)}} \\
Core  & \texttt{dem\_par} (positive rate) $\downarrow$          & 0.09938 & 0.06895 \\
Core  & \texttt{eq\_opp} (TPR) $\downarrow$                     & 0.05208 & 0.04351 \\
Other & \texttt{pred\_eq} (FPR) $\downarrow$                    & 0.51667 & 0.20000 \\
Other & \texttt{pred\_par} (PPV) $\downarrow$                   & 0.01265 & 0.00818 \\
Other & \texttt{forp} (FOR) $\downarrow$                        & 0.13971 & 0.10714 \\
Other & \texttt{acc\_eq} (accuracy) $\downarrow$                & 0.02461 & 0.04006 \\
Other & \texttt{f1\_score\_eq} (F1) $\downarrow$                & 0.01984 & 0.02542 \\
\midrule

\multicolumn{4}{l}{\textbf{Parallel audit (worst-case; violation $\downarrow$)}} \\
Core  & \texttt{dem\_par} (positive rate) $\downarrow$          & 0.09938 & 0.20435 \\
Core  & \texttt{eq\_opp} (TPR) $\downarrow$                     & 0.05208 & 0.09204 \\
Other & \texttt{pred\_eq} (FPR) $\downarrow$                    & 0.51667 & 0.60000 \\
Other & \texttt{pred\_par} (PPV) $\downarrow$                   & 0.07956 & 0.06273 \\
Other & \texttt{forp} (FOR) $\downarrow$                        & 0.39216 & 0.38889 \\
Other & \texttt{acc\_eq} (accuracy) $\downarrow$                & 0.05042 & 0.07769 \\
Other & \texttt{f1\_score\_eq} (F1) $\downarrow$                & 0.03792 & 0.06290 \\
\midrule

\multicolumn{4}{l}{\textbf{Intersectional audit (violation $\downarrow$)}} \\
Core  & \texttt{dem\_par} (positive rate) $\downarrow$          & 0.39007 & 0.41497 \\
Core  & \texttt{eq\_opp} (TPR) $\downarrow$                     & 0.24548 & 0.27363 \\
Other & \texttt{pred\_eq} (FPR) $\downarrow$                    & 1.16667 & 1.00000 \\
Other & \texttt{pred\_par} (PPV) $\downarrow$                   & 0.12558 & 0.10868 \\
Other & \texttt{forp} (FOR) $\downarrow$                        & 1.00000 & 1.08333 \\
Other & \texttt{acc\_eq} (accuracy) $\downarrow$                & 0.09790 & 0.13784 \\
Other & \texttt{f1\_score\_eq} (F1) $\downarrow$                & 0.11008 & 0.12537 \\
\bottomrule
\end{tabular}

\caption{Full ABCFair metrics for two prompting modes on \texttt{ACSPublicCoverage}. Arrows indicate the preferred direction (\(\uparrow\) higher is better; \(\downarrow\) lower is better). Fairness quantities are ABCFair group-disparity violations (lower is better) under three auditing granularities: \emph{binary}, \emph{parallel} (worst-case over sensitive attributes), and \emph{intersectional}.}
\label{tab:abcfair_full_exp_values}
\end{table*}

\textbf{Additional metrics explanation.} In addition to the two \emph{core} notions emphasized above, demographic parity and equal opportunity, we also log five additional notions used in the benchmark: predicted equality (FPR parity), predictive parity (PPV parity), false-omission-rate parity (FOR parity), accuracy equality, and F1-score equality (all reported as violations; lower is better).

\textbf{Spillover effects and fairness granularity.} In our fairness prompt, we explicitly specify protecting the \texttt{gender} feature. However, the \texttt{SchoolPerformance} dataset contains a second protected feature: \texttt{parental\_education}. We choose not to disclose this attribute to test spillover effects. Specifically, ABCFair defines violations not only for single features but also across groups of features as: \textbf{binary} (a single sensitive feature binarized into advantaged/disadvantaged), \textbf{parallel} (multiple sensitive features audited independently with worst-case aggregation), and \textbf{intersectional} (groups defined by intersections of sensitive features)~\cite{defrance2024abcfair}.

Table~\ref{tab:abcfair_full_exp_values} reports the complete ABCFair metric log for our two prompting modes on \texttt{SchoolPerformance}: \textbf{Simple Talking Tree} (default prompt) and \textbf{Fair Talking Tree} (fairness prompt). Our model successfully reduces gender bias, the protected attribute mentioned in the prompt. However, we also observe increased bias with respect to \texttt{parental\_education}, which becomes visible in the \textbf{parallel} and \textbf{intersectional} evaluations. This highlights an important cautionary lesson: protected features should be specified from the start, and spillover effects should be audited explicitly.

\textbf{Takeaways.}
These results demonstrate that prompt-level fairness interventions are effective and targeted. When we instruct the model to protect a specific attribute, it does so successfully. However, practitioners must carefully specify \emph{all} relevant protected attributes in their prompts; fairness improvements do not automatically generalize to unmentioned sensitive features. We view these ABCFair results as evidence of \emph{controllability}: our pipeline can be steered via natural language toward specific fairness objectives, and the resulting behavior can be transparently audited using standardized reporting.

\section{Monotonicity Experimental Details}\label{app:monotonicity}

Monotonicity is a fundamental property of machine learning models that encourages predictions to align with common sense and domain-specific expert logic.
Formally, a model \( f: \mathbb{R}^d \rightarrow \mathbb{R} \) is \emph{monotonically nondecreasing} with respect to feature \( i \) if for any two instances
\( \mathbf{x}, \mathbf{x}' \in \mathcal{X} \) such that \( x_i \leq x'_i \) and all other features are equal, it holds that
\( f(\mathbf{x}) \leq f(\mathbf{x}') \).
Monotonically decreasing behavior is defined analogously.
This property is especially important in applied tasks where the direction of a feature's influence is known \emph{a priori}: for example, in credit scoring, the probability of default is logically expected to increase with a client's debt burden.

In our experiments, we attempted to build trees monotonic with respect to specified features.
For GPT-5-generated trees, we used a single prompt explicitly requiring the model to take the given monotonicity constraints into account during tree construction.

\textbf{Monotonicity definition:}
A model \( f(\mathbf{x}) \) is monotonically nondecreasing with respect to the \( i \)-th feature if for any two instances \( \mathbf{x} \) and \( \mathbf{x}' \) such that \( x_i \leq x'_i \) and all other features are equal, it holds that \( f(\mathbf{x}) \leq f(\mathbf{x}') \). Monotonically decreasing behavior is defined analogously with the opposite inequality.

\textbf{ICE-based NMI (Normalized Monotonicity Index):}
To evaluate monotonicity, we use Individual Conditional Expectation (ICE) analysis~\cite{Goldstein02012015}. For each test instance, we vary the constrained feature from its minimum to maximum value while keeping all other features fixed at their original values, and observe how the model's prediction changes.

For each test row and each constrained feature, we examine whether the prediction moves in the required direction as the feature increases. For monotonically increasing constraint: whenever the feature value increases, the prediction should NOT decrease. Each decrease counts as a violation within that row. For monotonically decreasing constraint: whenever the feature value increases, the prediction should NOT increase. Each increase counts as a violation within that row.

Let \( B \) be the number of test rows where no violation occurs (the model's predictions move correctly or stay flat throughout the entire ICE curve), and let \( A \) be the number of test rows where at least one violation occurs (there is at least one local non-monotonic segment). Then NMI is calculated as:
\[
\mathrm{NMI} = \frac{B - A}{B + A}
\]

NMI ranges from:
\begin{itemize}
    \item \( +1 \): perfect monotonicity (all rows satisfy the constraint, \( B = \text{total rows} \), \( A = 0 \))
    \item \( 0 \): violations and non-violations occur equally often (\( B = A \))
    \item \( -1 \): all rows contain at least one violation (\( A = \text{total rows} \), \( B = 0 \))
\end{itemize}

\textbf{Monotonicity constraints (respect these when building the tree):}

The exact prompt provided to GPT-5 was as follows:
\begin{tcolorbox}[
    colframe=black!70,
    colback=gray!5,
    title=Monotonicity prompt,
    enhanced,
    breakable,
    skin first=enhanced,
    skin middle=enhanced,
    skin last=enhanced,]
\small\ttfamily
Your tree will be evaluated on NMI = (B - A) / (B + A), 
where B is the number of test rows on which the model's predictions move in the correct direction as the constrained feature increases, and A is the number of rows where at least one violation occurs. 
NMI ranges from $-1$ (all rows violate) to $+1$ (no violations). Maximise NMI. Monotonicity constraints (respect these when building the tree):
    
[For each dataset, the constraints were specified as follows:]
\begin{itemize}
    \item For monotonically increasing features: "As the following features increase, the target must also increase: \{feature names\}."
    \item For monotonically decreasing features: "As the following features increase, the target must decrease: \{feature names\}."
\end{itemize}
\end{tcolorbox}

As a baseline, we compared these trees with conventional CART models trained without monotonicity constraints, as well as with LightGBM models configured with monotonic constraints. LightGBM enforces monotonicity by design when monotonic constraints are specified, so its NMI is always perfect by construction.

We compare CART, GPT-5-generated trees, and LightGBM.
For LightGBM, we consider the default ensemble size with \( n_{\mathrm{e}}=100 \) estimators and a single-estimator version with \( n_{\mathrm{e}}=1 \).

\textbf{Evaluation protocol for NMI:}
We estimate NMI via Monte Carlo sampling: for each test instance and constrained feature, we sample 100 feature values and check for monotonicity violations. This is repeated for 100 test instances. For models with multiple constrained features, we report the average NMI across all such features.




\section{Data leak}\label{app:data_leak}

Designing a convincing leakage test for an {agentic} LLM is surprisingly subtle. A seemingly clean solution is to benchmark on purely synthetic data, but this is ill-posed for our setting: 1) it is unclear how to construct synthetic generators whose method rankings transfer to real-world datasets, and 2) synthetic features lack semantic meaning, whereas our agent is explicitly intended to exploit broad world knowledge (e.g., plausible feature relationships, reasonable preprocessing choices, and inductive biases suggested by informal metadata). Consequently, a synthetic-only evaluation would likely conflate the downsides of LLM-guided hypothesis generation with the data leakage.

Instead, we use a two-axis stress test that preserves the real evaluation task while selectively removing the information that memorization-based shortcuts would rely on. We always evaluate on the held-out {clean} test set, and progressively corrupt only the train-validation environment along 1) \textbf{data corruption} and 2) \textbf{metadata corruption}. Data corruption is controlled by four discrete noise levels: {Clean}; {Mild} (small numeric perturbations via random rotations of a subset of numeric features, per-feature scaling, additive noise, and sparse masking); {Some} (shuffling values within 50\% of columns); and {Complete} (shuffling all columns and permuting the training labels). Metadata corruption is controlled by \texttt{metadata\_mode}: we provide either the true dataset description, no description, or a misleading description copied from another dataset. We additionally report two non-LLM baselines trained under the same corruptions: \textbf{CatBoost}, as a strong off-the-shelf tabular model, and a \textbf{naive} estimator that ignores features (class-frequency predictor for classification; mean target for regression).

A critical nuance is the {stealthiness} of the complete corruption: column-wise shuffling preserves printed values and univariate (marginal) statistics while destroying the joint dependence structure that carries predictive signal. Since the agent typically inspects only small data snippets and summary statistics, the corrupted train set remains superficially plausible and does not trivially ``announce'' itself as synthetic. Under our definition, ``cheating'' means that the LLM recognizes the dataset as a known benchmark (e.g., via its name, feature distribution signatures, or remembered community solutions) and then attempts to construct a model that matches memorized test-set behavior or a memorized high-performing solution, rather than fitting to the provided labels. If such recognition-and-retrieval were occurring, then performance on the clean test set would be relatively insensitive to removing the training signal, especially in the setting with the {True metadata} but {Complete} train-validation corruption, where the dataset description and marginal statistics remain intact.

\begin{table*}[t]
\centering
\caption{Evaluating agent-constructed decision trees and advanced tabular methods on low-resource tabular datasets. The datasets are split into three groups by task type. \textcolor{gray}{Gray} values are adjusted standard deviations over 5 splits (TabArena v0.1).}
\label{tab:full_data_leak}
\small
\setlength{\tabcolsep}{2pt}
\renewcommand{\arraystretch}{1.15}
\begin{adjustbox}{max width=\textwidth}

\begin{tabular}{lccccccccccccccc}
\toprule
\multirow{2}{*}{\makecell[l]{Noise}} &
\multicolumn{3}{c}{\makecell{True metadata}} &
\multicolumn{3}{c}{\makecell{No metadata}} &
\multicolumn{3}{c}{\makecell{Wrong metadata}} &
\multicolumn{3}{c}{\makecell{CatBoost}} &
\multicolumn{3}{c}{\makecell{Baseline}} \\
\cmidrule(lr){2-4}\cmidrule(lr){5-7}\cmidrule(lr){8-10}\cmidrule(lr){11-13}\cmidrule(lr){14-16}
&
\makecell{Fitness\\{\tiny\textcolor{gray}{ROC AUC$\uparrow$}}} &
\makecell{Anneal\\{\tiny\textcolor{gray}{LogLoss$\downarrow$}}} &
\makecell{AirfoilSN\\{\tiny\textcolor{gray}{RMSE$\downarrow$}}} &
\makecell{Fitness\\{\tiny\textcolor{gray}{ROC AUC$\uparrow$}}} &
\makecell{Anneal\\{\tiny\textcolor{gray}{LogLoss$\downarrow$}}} &
\makecell{AirfoilSN\\{\tiny\textcolor{gray}{RMSE$\downarrow$}}} &
\makecell{Fitness\\{\tiny\textcolor{gray}{ROC AUC$\uparrow$}}} &
\makecell{Anneal\\{\tiny\textcolor{gray}{LogLoss$\downarrow$}}} &
\makecell{AirfoilSN\\{\tiny\textcolor{gray}{RMSE$\downarrow$}}} &
\makecell{Fitness\\{\tiny\textcolor{gray}{ROC AUC$\uparrow$}}} &
\makecell{Anneal\\{\tiny\textcolor{gray}{LogLoss$\downarrow$}}} &
\makecell{AirfoilSN\\{\tiny\textcolor{gray}{RMSE$\downarrow$}}} &
\makecell{Fitness\\{\tiny\textcolor{gray}{ROC AUC$\uparrow$}}} &
\makecell{Anneal\\{\tiny\textcolor{gray}{LogLoss$\downarrow$}}} &
\makecell{AirfoilSN\\{\tiny\textcolor{gray}{RMSE$\downarrow$}}} \\
\midrule

Clean &
\shortstack{\(0.800\)\\{\tiny\textcolor{gray}{\(\pm 0.022\)}}} &
\shortstack{\(0.176\)\\{\tiny\textcolor{gray}{\(\pm 0.073\)}}} &
\shortstack{\(3.621\)\\{\tiny\textcolor{gray}{\(\pm 0.335\)}}} &
\shortstack{\(0.813\)\\{\tiny\textcolor{gray}{\(\pm 0.020\)}}} &
\shortstack{\(0.188\)\\{\tiny\textcolor{gray}{\(\pm 0.071\)}}} &
\shortstack{\(3.231\)\\{\tiny\textcolor{gray}{\(\pm 0.090\)}}} &
\shortstack{\(0.803\)\\{\tiny\textcolor{gray}{\(\pm 0.016\)}}} &
\shortstack{\(0.162\)\\{\tiny\textcolor{gray}{\(\pm 0.020\)}}} &
\shortstack{\(3.851\)\\{\tiny\textcolor{gray}{\(\pm 0.586\)}}} &
\shortstack{\(0.823\)\\{\tiny\textcolor{gray}{\(\pm 0.023\)}}} &
\shortstack{\(0.075\)\\{\tiny\textcolor{gray}{\(\pm 0.044\)}}} &
\shortstack{\(1.767\)\\{\tiny\textcolor{gray}{\(\pm 0.106\)}}} &
\shortstack{\(0.500\)\\{\tiny\textcolor{gray}{\(\pm 0.000\)}}} &
\shortstack{\(0.831\)\\{\tiny\textcolor{gray}{\(\pm 0.001\)}}} &
\shortstack{\(6.907\)\\{\tiny\textcolor{gray}{\(\pm 0.069\)}}} \\

Mild &
\shortstack{\(0.787\)\\{\tiny\textcolor{gray}{\(\pm 0.000\)}}} &
\shortstack{\(0.411\)\\{\tiny\textcolor{gray}{\(\pm 0.029\)}}} &
\shortstack{\(7.732\)\\{\tiny\textcolor{gray}{\(\pm 5.139\)}}} &
\shortstack{\(0.795\)\\{\tiny\textcolor{gray}{\(\pm 0.000\)}}} &
\shortstack{\(0.886\)\\{\tiny\textcolor{gray}{\(\pm 0.147\)}}} &
\shortstack{\(8.041\)\\{\tiny\textcolor{gray}{\(\pm 4.705\)}}} &
\shortstack{\(0.798\)\\{\tiny\textcolor{gray}{\(\pm 0.000\)}}} &
\shortstack{\(0.524\)\\{\tiny\textcolor{gray}{\(\pm 0.005\)}}} &
\shortstack{\(7.718\)\\{\tiny\textcolor{gray}{\(\pm 4.317\)}}} &
\shortstack{\(0.819\)\\{\tiny\textcolor{gray}{\(\pm 0.021\)}}} &
\shortstack{\(0.251\)\\{\tiny\textcolor{gray}{\(\pm 0.002\)}}} &
\shortstack{\(7.341\)\\{\tiny\textcolor{gray}{\(\pm 4.480\)}}} &
\shortstack{\(0.500\)\\{\tiny\textcolor{gray}{\(\pm 0.000\)}}} &
\shortstack{\(0.831\)\\{\tiny\textcolor{gray}{\(\pm 0.001\)}}} &
\shortstack{\(6.907\)\\{\tiny\textcolor{gray}{\(\pm 0.069\)}}} \\

Some &
\shortstack{\(0.743\)\\{\tiny\textcolor{gray}{\(\pm 0.066\)}}} &
\shortstack{\(0.471\)\\{\tiny\textcolor{gray}{\(\pm 0.021\)}}} &
\shortstack{\(5.170\)\\{\tiny\textcolor{gray}{\(\pm 2.019\)}}} &
\shortstack{\(0.743\)\\{\tiny\textcolor{gray}{\(\pm 0.066\)}}} &
\shortstack{\(0.346\)\\{\tiny\textcolor{gray}{\(\pm 0.032\)}}} &
\shortstack{\(5.239\)\\{\tiny\textcolor{gray}{\(\pm 1.876\)}}} &
\shortstack{\(0.738\)\\{\tiny\textcolor{gray}{\(\pm 0.060\)}}} &
\shortstack{\(0.325\)\\{\tiny\textcolor{gray}{\(\pm 0.019\)}}} &
\shortstack{\(4.972\)\\{\tiny\textcolor{gray}{\(\pm 1.929\)}}} &
\shortstack{\(0.744\)\\{\tiny\textcolor{gray}{\(\pm 0.082\)}}} &
\shortstack{\(0.249\)\\{\tiny\textcolor{gray}{\(\pm 0.017\)}}} &
\shortstack{\(4.249\)\\{\tiny\textcolor{gray}{\(\pm 1.801\)}}} &
\shortstack{\(0.500\)\\{\tiny\textcolor{gray}{\(\pm 0.000\)}}} &
\shortstack{\(0.831\)\\{\tiny\textcolor{gray}{\(\pm 0.001\)}}} &
\shortstack{\(6.907\)\\{\tiny\textcolor{gray}{\(\pm 0.069\)}}} \\

Complete &
\shortstack{\(0.515\)\\{\tiny\textcolor{gray}{\(\pm 0.050\)}}} &
\shortstack{\(0.845\)\\{\tiny\textcolor{gray}{\(\pm 0.013\)}}} &
\shortstack{\(7.216\)\\{\tiny\textcolor{gray}{\(\pm 0.228\)}}} &
\shortstack{\(0.495\)\\{\tiny\textcolor{gray}{\(\pm 0.075\)}}} &
\shortstack{\(0.839\)\\{\tiny\textcolor{gray}{\(\pm 0.007\)}}} &
\shortstack{\(7.209\)\\{\tiny\textcolor{gray}{\(\pm 0.071\)}}} &
\shortstack{\(0.492\)\\{\tiny\textcolor{gray}{\(\pm 0.050\)}}} &
\shortstack{\(1.013\)\\{\tiny\textcolor{gray}{\(\pm 0.121\)}}} &
\shortstack{\(7.067\)\\{\tiny\textcolor{gray}{\(\pm 0.246\)}}} &
\shortstack{\(0.450\)\\{\tiny\textcolor{gray}{\(\pm 0.015\)}}} &
\shortstack{\(0.828\)\\{\tiny\textcolor{gray}{\(\pm 0.003\)}}} &
\shortstack{\(6.966\)\\{\tiny\textcolor{gray}{\(\pm 0.052\)}}} &
\shortstack{\(0.500\)\\{\tiny\textcolor{gray}{\(\pm 0.000\)}}} &
\shortstack{\(0.831\)\\{\tiny\textcolor{gray}{\(\pm 0.001\)}}} &
\shortstack{\(6.907\)\\{\tiny\textcolor{gray}{\(\pm 0.069\)}}} \\

\bottomrule
\end{tabular}

\end{adjustbox}
\end{table*}

Empirically, Table~\ref{tab:full_data_leak} shows the opposite: once the dependence structure is destroyed, performance collapses to near-baseline levels across tasks and metadata modes. For the agent-constructed trees, {Fitness} drops from approximately $0.800$ ROC AUC in the Clean setting (e.g., $0.800$ with True metadata; $0.813$ with No metadata; $0.803$ with Wrong metadata) to $0.49$--$0.52$ under Complete corruption ($0.515$, $0.495$, $0.492$), close to the naive $0.500$ baseline. On {Airfoil}, RMSE degrades from $ 3.2$--$3.9$ (Clean) to $ 7.07$--$7.22$ (Complete), approaching the naive baseline of $6.907$. On {Anneal}, LogLoss worsens from $ 0.16$--$0.19$ (Clean) to $ 0.84$--$1.01$ (Complete), again near the baseline region (e.g., $0.831$ for the soft class-frequency baseline). CatBoost exhibits a similar pattern under the same corruptions (e.g., Fitness drops from $0.823$ to $0.450$ under Complete; Airfoil rises from $1.767$ to $6.966$; Anneal rises from $0.075$ to $0.828$), supporting the interpretation that the corruption truly removes learnable signal rather than merely confusing the agent. Conversely, when the data remain informative, metadata can matter in a dataset-dependent way: for example, on Clean {Airfoil} the agent improves with No metadata ($3.231$) relative to True metadata ($3.621$) and degrades with Wrong metadata ($3.851$), indicating that semantic priors can help or hurt depending on how well they align with the empirical signal; however, these differences vanish once the signal is eliminated (all metadata modes converge near $ 7.1$ RMSE under Complete).

We emphasize that this test is not perfect: because the model is agentic, it could in principle detect subtle inconsistencies and decide not to ``apply'' a memorized solution when it suspects the dataset is not the canonical benchmark. This is precisely why we favor dependence-destroying but marginal-preserving corruptions, and why we leave the dataset name/description unchanged in the strongest condition: the agent continues to see the same high-level framing and superficially plausible marginals, so any recognition-based shortcut should still be triggered. Within this design, the consistent collapse under Complete corruption provides evidence that our reported gains are tethered to the provided training signal rather than memorized benchmark artifacts.

Finally, our use of ``cheating'' focuses on dataset recognition followed by retrieval of memorized labels/solutions. Other leakage pathways that we do {not} directly isolate here include 1) memorization of {preprocessing pipelines} that happen to work well on a dataset family or 2) heuristic overfitting to the validation split through extensive trial-and-error. Overall, the evidence in Table~\ref{tab:full_data_leak} indicates that, under a strong and relatively stealthy corruption regime, both the LLM agent and CatBoost fail when the training dependence structure is removed, supporting the conclusion that our main results are not driven by pre-training memorization of the test set.

\section{Impact Statement}
\label{sec:limitations}

This work aims to make machine learning more accessible and interpretable by enabling domain experts to guide model construction through natural language. The resulting decision trees are inherently more transparent than black-box alternatives, which may benefit high-stakes domains where explainability is required for compliance and trust.

However, this accessibility raises ethical considerations. Natural language instructions could encode human biases; for instance, domain experts might inadvertently (or deliberately) specify criteria that disadvantage protected groups. While the interpretable tree structure makes such biases easier to audit than in opaque models, practitioners should carefully review both the instructions and the resulting trees.

By lowering the barrier to model construction, this work could democratize machine learning, enabling domain experts without ML expertise to build predictive models. While this has clear benefits for accessibility, it also risks proliferating models built without proper validation or consideration of failure modes, potentially leading to poor decisions in critical applications. We recommend that organizations establish review processes for LLM-generated models before deployment.

Deploying this framework with cloud-based LLMs requires transmitting potentially sensitive tabular data to external providers, raising privacy concerns. While local or open-source models offer alternatives (see Section~\ref{sect:exp_ablation}), practitioners should carefully evaluate data governance policies before using commercial LLM APIs with confidential datasets, particularly in regulated industries such as healthcare or finance.

From a safety perspective, LLM agents with free-form communication capabilities pose risks in high-stakes settings~\cite{he2024emerged}. These can be mitigated by air-gapping the system~\cite{ruan2023_toolemu,bagdasarian2024_airgapagent} --- running Python in isolated containers and serializing trees in formats that disallow arbitrary code execution. Inspecting reasoning traces~\cite{korbak2025_cot_monitorability} offers another safeguard, though the absence of problematic reasoning is not a guarantee of safe behavior~\cite{turpin2023_unfaithful_cot}.

Finally, our experiments use proprietary GPT-5, which may affect reproducibility as the model evolves. Section~\ref{sect:exp_ablation} demonstrates that open-source LLMs can also leverage the framework, though with reduced performance.

\newpage
\section*{NeurIPS Paper Checklist}

\begin{enumerate}

\item {\bf Claims}
    \item[] Question: Do the main claims made in the abstract and introduction accurately reflect the paper's contributions and scope?
    \item[] Answer: \answerYes{} 
    \item[] Justification: The abstract and introduction accurately reflect the results obtaind in the paper. 

\item {\bf Limitations}
    \item[] Question: Does the paper discuss the limitations of the work performed by the authors?
    \item[] Answer: \answerYes{} 
    \item[] Justification: We discuss limitations of our method throughout the work and provide a summary of the impact in Appendix~\ref{sec:limitations}.

\item {\bf Theory assumptions and proofs}
    \item[] Question: For each theoretical result, does the paper provide the full set of assumptions and a complete (and correct) proof?
    \item[] Answer: \answerNA{} 
    \item[] Justification: The paer is methodological with the main claims supported by the empirical evidence. 

    \item {\bf Experimental result reproducibility}
    \item[] Question: Does the paper fully disclose all the information needed to reproduce the main experimental results of the paper to the extent that it affects the main claims and/or conclusions of the paper (regardless of whether the code and data are provided or not)?
    \item[] Answer: \answerYes{} 
    \item[] Justification: We document our experimental setup, including dataset used, model hyperparameters, tuning configurations in Appendix~\ref{app:eval_setup}. We also provide the code as part of the submission.

\item {\bf Open access to data and code}
    \item[] Question: Does the paper provide open access to the data and code, with sufficient instructions to faithfully reproduce the main experimental results, as described in supplemental material?
    \item[] Answer: \answerYes{} 
    \item[] Justification: We provide the code as part of the submission.

\item {\bf Experimental setting/details}
    \item[] Question: Does the paper specify all the training and test details (e.g., data splits, hyperparameters, how they were chosen, type of optimizer) necessary to understand the results?
    \item[] Answer: \answerYes{} 
    \item[] Justification: We provide the details of our experiments in Appendix~\ref{app:eval_setup}.

\item {\bf Experiment statistical significance}
    \item[] Question: Does the paper report error bars suitably and correctly defined or other appropriate information about the statistical significance of the experiments?
    \item[] Answer: \answerYes{} 
    \item[] Justification: We provide standard deviations across random splits of the data and model re-runs.

\item {\bf Experiments compute resources}
    \item[] Question: For each experiment, does the paper provide sufficient information on the computer resources (type of compute workers, memory, time of execution) needed to reproduce the experiments?
    \item[] Answer: \answerYes{} 
    \item[] Justification: We provide the details of the resources in Appendix~\ref{app:eval_setup}.
    
\item {\bf Code of ethics}
    \item[] Question: Does the research conducted in the paper conform, in every respect, with the NeurIPS Code of Ethics \url{https://neurips.cc/public/EthicsGuidelines}?
    \item[] Answer: \answerYes{} 
    \item[] Justification: The research in the paper adheres to the code of ethics. 

\item {\bf Broader impacts}
    \item[] Question: Does the paper discuss both potential positive societal impacts and negative societal impacts of the work performed?
    \item[] Answer: \answerYes{} 
    \item[] Justification: We provide the impact statement in Appendix~\ref{sec:limitations}.
    
\item {\bf Safeguards}
    \item[] Question: Does the paper describe safeguards that have been put in place for responsible release of data or models that have a high risk for misuse (e.g., pre-trained language models, image generators, or scraped datasets)?
    \item[] Answer: \answerNA{} 
    \item[] Justification: We do not release any pre-trained models or re-usable artifacts as part of the paper. 

\item {\bf Licenses for existing assets}
    \item[] Question: Are the creators or original owners of assets (e.g., code, data, models), used in the paper, properly credited and are the license and terms of use explicitly mentioned and properly respected?
    \item[] Answer: \answerYes{} 
    \item[] Justification: We cite the sources for all our models and datasets. 

\item {\bf New assets}
    \item[] Question: Are new assets introduced in the paper well documented and is the documentation provided alongside the assets?
    \item[] Answer: \answerNA{} 
    \item[] Justification: We do not ship any new assets as part of the paper. 

\item {\bf Crowdsourcing and research with human subjects}
    \item[] Question: For crowdsourcing experiments and research with human subjects, does the paper include the full text of instructions given to participants and screenshots, if applicable, as well as details about compensation (if any)? 
    \item[] Answer: \answerNA{} 
    \item[] Justification: The paper research did not involve crowdsourcing. 

\item {\bf Institutional review board (IRB) approvals or equivalent for research with human subjects}
    \item[] Question: Does the paper describe potential risks incurred by study participants, whether such risks were disclosed to the subjects, and whether Institutional Review Board (IRB) approvals (or an equivalent approval/review based on the requirements of your country or institution) were obtained?
    \item[] Answer: \answerNA{} 
    \item[] Justification: Our study did not involve human subjects. 

\item {\bf Declaration of LLM usage}
    \item[] Question: Does the paper describe the usage of LLMs if it is an important, original, or non-standard component of the core methods in this research? Note that if the LLM is used only for writing, editing, or formatting purposes and does \emph{not} impact the core methodology, scientific rigor, or originality of the research, declaration is not required.
    \item[] Answer: \answerYes{} 
    \item[] Justification: Our main method relies on the LLMs. We discuss the potential shortcomings and risks in Appendix~\ref{sec:limitations}.

\end{enumerate}

\end{document}